# Beyond Binary Rooftop Mapping: A Four-Class Deep Learning Framework for Green Roof Potential Assessment from Open Swiss Geospatial Data

Htet Yamin Ko Ko*

*Abstract*—**The development of effective urban climate adaptation strategies requires comprehensive spatial information on rooftops and buildings, since such information underpins the assessment of ecosystem services provided by green infrastructure, particularly for urban heat island (UHI) mitigation. Although green roofs are widely acknowledged as a promising measure for improving urban thermal comfort, most existing research maps either current green rooftops or rooftops with greening potential, but not both. This study presents a modified deep convolutional neural network rooftop classification framework based on Roofpedia, developed by the Urban Analytics Lab at the National University of Singapore. The proposed model combines high resolution aerial imagery with rooftop slope information derived from a digital surface model and relies entirely on publicly available Swisstopo datasets: SWISSIMAGE orthophotos, swissSURFACE3D elevation data, and swissTLM3D building footprints. Applied to Bern, Switzerland, the model labels rooftops into four categories: existing green roofs, rooftops suitable for green roof installation, rooftops with solar panels, and flat rooftops unsuitable for greening. Slope thresholds are integrated alongside spectral analysis to reflect real world installation conditions. On the four-class task the model achieved a training mean Intersection over Union (mIoU) of 0.86 and a validation mIoU of 0.78. For the adopted SwissImage configuration, building-level classification reached 51% overall accuracy across all four classes, a substantially harder task than the binary green/non-green mapping of existing baselines (77.5% for Roofpedia on its single class), while delivering considerably richer, policy-relevant output. The framework identifies realistic opportunities for green roof expansion and supplies urban planners with evidence-based information for green infrastructure deployment in Bern and other Swiss cities. Because it is fully open source, the framework is transferable to cities worldwide.**

## I. INTRODUCTION

As urban areas expand and cities grow denser, innovative approaches are needed to accommodate green infrastructure within the built environment. Vertical, compact cities are widely regarded as more sustainable, and using available rooftop areas for civic spaces, solar energy generation, and urban farming has become an increasingly common strategy [1]. Usage of flat rooftop areas for other purposes are not a modern approach; it was implemented in traditional Egyptian houses to extend their living spaces [1]. Today, installing greenery on flat rooftop surfaces has become an essential element of modern, sustainable urban infrastructure [1].

Thus, accurate identification and classification of urban rooftops play a critical role in supporting climate adaptation strategies, renewable energy planning, and nature-based solutions [2]. Although green roofs and solar installations are increasingly promoted as climate adaptation and mitigation strategies, their spatial distribution and rooftop suitability remain insufficiently understood at the city scale. Existing studies on rooftop mapping in Switzerland focus almost exclusively on Zurich and Geneva, leaving major urban centers such as Bern understudied despite their national importance [3]. Existing datasets and models, such as Roofpedia [4], primarily focus on mapping existing green and solar rooftops, while other recent studies (e.g., the Swiss Territorial Data Lab [3]) classify existing roofs into different roof vegetation categories based on commercial aerial imagery. Similarly, several models have been developed to detect solar photovoltaic rooftop installation or assess rooftop solar potential [5], [6], [7]. However, current approaches remain limited in scope: they either identify existing features or evaluate solar suitability, but no existing model provides an integrated classification of rooftop surfaces into (i) existing green rooftops, (ii) potential green rooftops, (iii) existing solar rooftops, and (iv) flat roofs unsuitable for green infrastructure installation.

Moreover, some approaches rely on commercial high-resolution imagery, which limits reproducibility and accessibility [3]. Some researchers have conducted work in mapping Green Roofs by using satellite imagery, Geographic Information System (GIS) and machine learning technologies [8], [9], [10], [11], [12], [13], [14], [15]. Meanwhile, deep learning models and High Spatial Resolution (HSR) aerial images play a critical role in rooftop mapping [2], [16]. As the number of available deep learning pipelines and aerial image products continues to grow, systematic testing and comparison are needed to identify optimal solutions. Comparative evaluations of FCN, U-Net, and DeepLabv3+ on high-resolution aerial imagery found that architecture choice substantially affects both accuracy and efficiency of rooftop extraction, with encoder-decoder designs outperforming FCN [2]. These gaps highlight the need for an open-source, city-wide, multi-class rooftop classification framework capable of identifying realistic opportunities for green infrastructure

Htet Yamin Ko Ko (0000-0002-8242-0767) is with the International Space Science Institute, Bern, Switzerland (e-mail: koko.htetyamin@issibern.ch).
This work was supported by the ISSI Fellowship. *Corresponding author.

expansion in Bern. Without such a framework, cities remain limited in their ability to implement coordinated planning for urban cooling strategies.

To address this limitation, the present study extends and modifies the Roofpedia deep learning framework to deliver a comprehensive, multi-class rooftop classification model capable of supporting urban greening and energy transition initiatives.

This study introduces the first fully open-source, multi-class rooftop classification model that simultaneously distinguishes existing green roofs, existing solar rooftops, potential green rooftops, and non-suitable flat surfaces. By extending and modifying the Roofpedia deep learning architecture with area-based label assignment for multiclass U-Net segmentation and integrating open-source Digital Surface Model (DSM) to evaluate the elevation constraint, the proposed framework enables a more realistic and actionable assessment of green roof potential in urban areas. This approach fills a major methodological gap in rooftop mapping and provides a transparent, reproducible tool for supporting climate adaptation and sustainable urban planning.

The contribution of this paper is threefold. First, we introduce a four-class rooftop classification system for existing green roofs, existing solar roofs, potential green roofs, and non-suitable flat surfaces, which has not been achieved in previous studies. Second, the study implements and evaluates three deep learning models and compares them with respect to their accuracy and efficiency in rooftop extraction. Third, the study assesses the performance of three different aerial image sources with the proposed four-class rooftop classification pipeline.

Research Objectives are as follows:

- To develop a four-class rooftop classification scheme (existing green roofs, existing solar roofs, potential green roofs, and non-suitable flat surfaces) by modifying the Roofpedia deep learning model with slope-based suitability constraints derived from open-source Swisstopo Digital Surface Model.
- To implement and evaluate three deep learning models for rooftop extraction and classification, and to compare their accuracy and computational efficiency.
- To assess the performance of three different aerial image sources with the proposed four-class rooftop classification pipeline.
- To apply the best-performing pipeline to classify all rooftops in the City of Bern using a reproducible open-source workflow, and to generate the first open-source rooftop greening dataset for the city in support of green roofs as an ecosystem service for urban heat island mitigation.

## II. Background

With the intensifying effects of climate change and increasing temperatures, green rooftops become an important factor in urban planning, especially for the cities with dense urban density [3]. As the temperature is maxing every year and the increasing urban areas, the presence of green rooftops become an aspect for climate change mitigation measurement given the role they play in creating microclimate, enriching biodiversity and creating water retention [3]. Global assessments estimate that roof greening could lower daytime land surface temperatures in cities by roughly 0.6 to 1.6 degrees Celsius depending on the greening extent [17].

As many research studies exploited the ecosystem services provided by urban based green infrastructures, green rooftops become important for urban planning in the dense urban areas [3]. Green rooftops can provide thermal comfort, water retention, biodiversity support and aesthetic service [3].

Several machine learning and deep learning studies have been developed for the automatic rooftop mapping and potential green rooftops computation. [3] applied traditional machine learning and deep learning models to compare the performance of green rooftop mapping. The study utilized aerial images by the national aerial imagery survey (SWISSIMAGE RS), large-scale topographic landscape model of Switzerland swissTLM3D as the input data and applied a combined model of random forest and logistic regression and DeepLabV3 deep learning model. The study used Normalized Difference Vegetation Index (NDVI) as one of the feature bands to map the green rooftop.

In the literature, rooftops mapping can be performed by different methods applied on aerial imagery: thresholding on NDVI bands, classification by machine-learning-based on engineered features, and object detection by deep learning [2], [3]. When mapping rooftops, the geometries, slope, surface area, materials and depth of roofs play as decision factors [3], [8], [9]. The study utilized the high resolution of aerial photos with NIR band to classify the rooftops into six classes: Bare, Extensive, Spontaneous, Lawn, Intensive, and Terraces. Most of the studies only classified the rooftops into two classes (Green and Non-green rooftops) while this study classified the rooftops into six classes. The canopy height models (CHM) derived from LiDAR acquisitions have been used to mask pixels corresponding to the vegetation of overhanging trees [3].

Despite the growing interest in urban greening and rooftop-based climate adaptation measures, existing mapping approaches remain narrowly focused on single-purpose tasks. Current studies either identify existing green rooftops, classify vegetation types on rooftops, or detect existing/potential solar photovoltaic installations . However, none of these models integrate these components into a unified framework capable of distinguishing between rooftops that are already greened, rooftops occupied by solar systems, rooftops that are potentially suitable for green roof installation, and rooftops that are unsuitable. Moreover, most existing approaches either overlook structural and elevation constraints or rely on proprietary aerial imagery, limiting reproducibility and adoption by cities with fewer resources. Consequently, there is a critical need for an open, scalable, multi-class rooftop classification system that simultaneously accounts for rooftop condition, existing infrastructure, and structural suitability using fully open-source geospatial data.

The proposed study applied opensource national aerial imagery and land survey data from Swisstopo on mapping existing and potential green roofs and distinguishing among various roof types, including existing green rooftops, potential greening area, solar, and flat roofs which are not suitable to install green rooftops.

Deep learning and Geographic Information System are applied to develop potential and existing rooftop mapping models[16]. The results demonstrate substantial differences between classification approaches, highlighting the importance of integrating structural constraints in rooftop suitability assessments. The priority-based method identified the highest number of existing green roofs but also overestimated potential green surfaces due to its lack of physical feasibility checks. In contrast, the area- and slope-based methods produced identical results, suggesting that many rooftops in Bern fall within the feasible slope range for green roof installation. This reinforces the need for explicit structural criteria in automated assessments, as area-only approaches may misclassify steep or unsuitable rooftops as potential green surfaces.

The large number of rooftops categorized as potential green roofs indicates a significant untapped opportunity for climate mitigation and UHI reduction within Bern. Notably, the high prevalence of existing solar rooftops emphasizes the necessity to distinguish solar-dominated surfaces from greening opportunities, especially in cities prioritizing renewable energy. By integrating both elevation and contextual roof characteristics, the modified model provides a more realistic estimate of intervention potential than previous studies using only spectral or vegetation indicators.

Importantly, the reliance on open-source datasets ensures that the workflow can be reproduced, expanded to other Swiss cities, or adapted internationally. This strengthens the practical relevance of the research for public authorities and urban planners seeking low-cost, scalable methods for climate adaptation planning.

## III. Automated Rooftop Classification

### A. Study Area

Bern is the federal capital and fifth largest city of Switzerland, located at 46.94 N, 7.44 E on the Swiss Plateau (Mittelland) [18], [19]. The study area covers 156.7 km$^2$, the smallest study area among the major Swiss cities, encompassing the city of Bern and its neighboring municipalities. Bern has a population of approximately 135,000 with a population density of 2,673 inhabitants per km$^2$ [20]. The city has a temperate oceanic climate (Köppen Cfb) with an annual mean temperature of 10.3 degrees Celsius and annual precipitation of 1,021.8 mm [21]. The urban structure consists of a historic city center alongside dense residential areas in the old town region. The terrain is characterized by hills of up to 858 m and is crossed by the Aare River [22]. This varied topography and compact urban form make Bern a representative and challenging environment for rooftop-based green infrastructure assessment.

### B. Datasets

#### 1) SWISSIMAGE

The orthophoto mosaic SWISSIMAGE 10 cm is a composition of new digital color aerial photographs over the whole of Switzerland with a ground resolution of 10 cm in the plain areas and main alpine valleys and 25 cm over the Alps . It is updated in a cycle of 3 years. In this study, images from 2024 were used. The dataset is opensource [23].

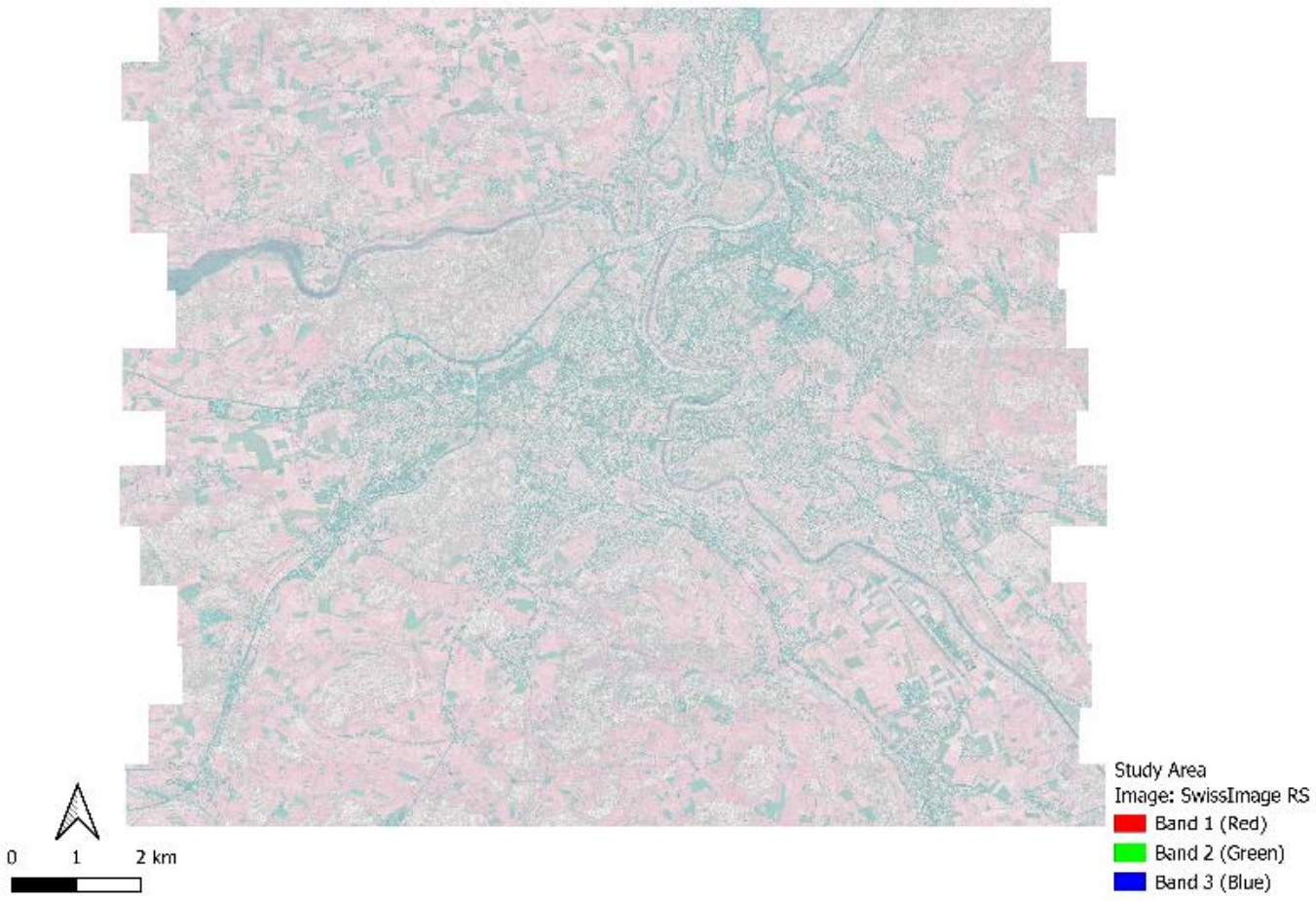

*Fig. 1: Study Area of Bern. The tiles are SwissImage RS (Red, Green, Blue and Near-infrared channels) dataset provided by Swiss Territorial Data Lab.*

#### 2) SWISSIMAGE RS [24]

SWISSIMAGE RS contains an unequalled wealth of information range for the years between 2005 and 2025. Bringing together the 4 spectral bands: near infrared, red, green and blue, these orthophotos provide basic data for many professional applications [24].

#### 3) swissTLM3D [25]

Data on building footprints have been extracted from swissTLM3D. swissTLM3D is the large-scale topographic landscape model of Switzerland and one of its most detailed vector datasets, comprising roads, buildings, watercourses, and land cover; building footprints were extracted from its building layer [25].

#### 4) swissSURFACE3D [26]

swissSURFACE3D Raster is a digital surface model (DSM) which represents the earth's surface including visible and permanent landscape elements such as soil, natural cover, and all sorts of constructive work with the exception of power lines and masts [26].

#### 5) Mapbox

The satellite imagery used in original Roofpedia pipeline is retrieved with the Mapbox static tiles API. According to Mapbox, the images available for access are from a combination of different sources, including Maxar's Vivid products for much of the world, Nearmap and USDA's NAIP 2011-2013 in the contiguous United States, and open aerial imagery from Denmark, Finland, Germany, and other regions. These datasets offer resolutions of about 50 cm and higher in major cities in the world.

*TABLE 1*

DETAILED INFORMATION OF DATASETS

| Dataset | Format | Opensource | Year | Resolution (meter) |
|---|---|---|---|---|
| SWISSIMAGE | Raster | Yes | 2024 | 0.1 |

| SWISSIMAGE RS | Raster | No | 2024 | 0.1 |
|---|---|---|---|---|
| swissTLM3D | Vector | Yes | 2025 | - |
| swissSURFACE3D | Raster | Yes | 2023 | 0.1 |
| Mapbox | Raster | Partial (rate-limited API) | - | 0.5 |

### C. Models

#### 1) Roofpedia

Roofpedia [4] is an open-source automated Python framework to provide a large-scale approach to map the existing green roofs and roofs with photovoltaic panels. The model is designed to work with a variety of imagery sources (Mapbox Tiles, Aerial Photographs from cities' open data repository) and building footprints dataset. A modified U-Net model with weights from ImageNet Dataset and ResNet50 backbone were used to map the existing green and solar rooftops across 17 cities. The original Roofpedia model can classify the roofs into two labels: green and solar. The code is available at https://github.com/ualsg/Roofpedia.

#### 2) Modified Roofpedia

Building upon and extending the Roofpedia framework, the proposed model classifies rooftops into four categories: existing green roofs, potential green roofs, solar-installed roofs, and surfaces unsuitable for greening. The model relies entirely on publicly available Swisstopo datasets, ensuring transparency, reproducibility, and long-term scalability. The code is available at https://github.com/StellaHtet92/Modified-Roofpedia.

#### 3) proj-vegroofs-DL

The proj-vegroofs-DL project was proposed by the Cantons of Zurich and Geneva and developed by the Swiss Territorial Data Lab, with project start in November 2023 [3]. The model is an adaptation of DeepLabV3, reusing its Atrous Spatial Pyramid Pooling (ASPP) module and processing its output for image classification . The model consists of three blocks: a convolutional encoder backbone with adjustable width and depth, the ASPP module, and a residual Multi-Layer Perceptron (MLP). It was trained for multiclass classification across six vegetation categories: bare, terrace, spontaneous, extensive, lawn, and intensive. Recall scores were 0.91, 0.77, 0.67, 0.79, 0.68, and 0.50 respectively [3]. And it also has binary class classification across bare vs vegetation roof categories. The code is available at https://github.com/swiss-territorial-data-lab/proj-vegroofs-DL.

### D. Metrics for evaluation

Model performance was monitored during training using three metrics. The cross-entropy loss was tracked to guide model optimization and detect overfitting. The mean Intersection over Union (mIoU) was used to measure the spatial agreement between predicted and reference rooftop classes, as it is a standard metric for semantic segmentation tasks [4]. In addition, Overall Accuracy was recorded to provide a general measure of the proportion of correctly classified pixels across all classes.

1) Loss (cross-entropy)

The cross-entropy loss was tracked throughout training as the primary optimization objective. It measures the difference between the predicted class probabilities and the reference labels at the pixel level. Monitoring the training and validation loss curves also allowed the detection of overfitting and informed decisions on early stopping and hyperparameter tuning.

$$\text{Cross Entropy} = -\sum_{n=1}^{N}\sum_{k=1}^{K}\sum_{j=1}^{P} d_{nkwh} \log p \quad (1)$$

Where K, N, and P are the number of classes, the mini-batch size, and the number of pixels [2]. p is the estimated probability for the class with reference label and d is one-hot encoding vector [2].

2) mIoU (mean Intersection over Union)

The mIoU was used as the main indicator of segmentation quality [4], [27]. It computes the overlap between predicted and reference regions for each class and averages the results across all classes. This metric is well suited to multi-class rooftop classification because it accounts for class imbalance and penalizes both false positives and false negatives.

$$\text{mIoU} = \frac{\sum_{k=1}^{K}\left[\frac{TP_k}{TP_k+FP_k+FN_k}\right]}{K} \quad (2)$$

where K is the number of classes, FP stands for false positive that represents the error between the reference label and prediction output where the prediction indicates the presence of a condition; TP stands for true positive when the model can provide the same prediction as the reference label; and FN stands for false negative[2].

3) Overall Accuracy (OA)

Overall Accuracy was recorded to provide a general measure of classification performance [3]. It represents the proportion of correctly classified pixels across all classes. While it is less sensitive to minority classes than mIoU, it offers an intuitive summary of model performance and enables comparison with previous rooftop mapping studies.

$$OA = \frac{TP+TN}{TP+TN+FP+FN} \quad (3)$$

where TP, FP and FN stand the same meaning as presented in EQUATION (2) and TN stands for True Negative when the model predicted the negative label correctly in the prediction [27].

### E. Dataset Labeling

The building dataset from Swisstopo is utilized as the baseline individual building footprint [28]. All buildings were manually labelled as four classes for Modified Roofpedia model, two classes for Roofpedia model, and two classes for proj-vegroofs-DL model.

### F. Study Workflow of Modified Roofpedia

The end-to-end processing workflow consists of four stages, as illustrated in Fig. 2.

**Stage 1: Input Data Preparation.** Four datasets are required as inputs as presented in TABLE 1. SwissImage was selected as the input source after benchmarking the Modified Roofpedia model with three candidate datasets: Mapbox, SwissImage, and SwissImage RS (Section IV-A). Mapbox and SwissImage achieved similar segmentation performance, whereas SwissImage RS suffered from reduced image sharpness at building scale. However, Mapbox is only partially open source, since its static tiles API impose a monthly limit on the number of retrievable tiles, which makes it impractical for larger study areas. SwissImage, in contrast, is fully open and unrestricted, and was therefore adopted for the city scale application. A total of 9,814 buildings were manually collected and labelled as training samples. The swissTLM3D building footprint dataset provides 103,074 vector geometries covering the full study area. swissSURFACE3D 2.0 provides the digital surface model for slope computation.

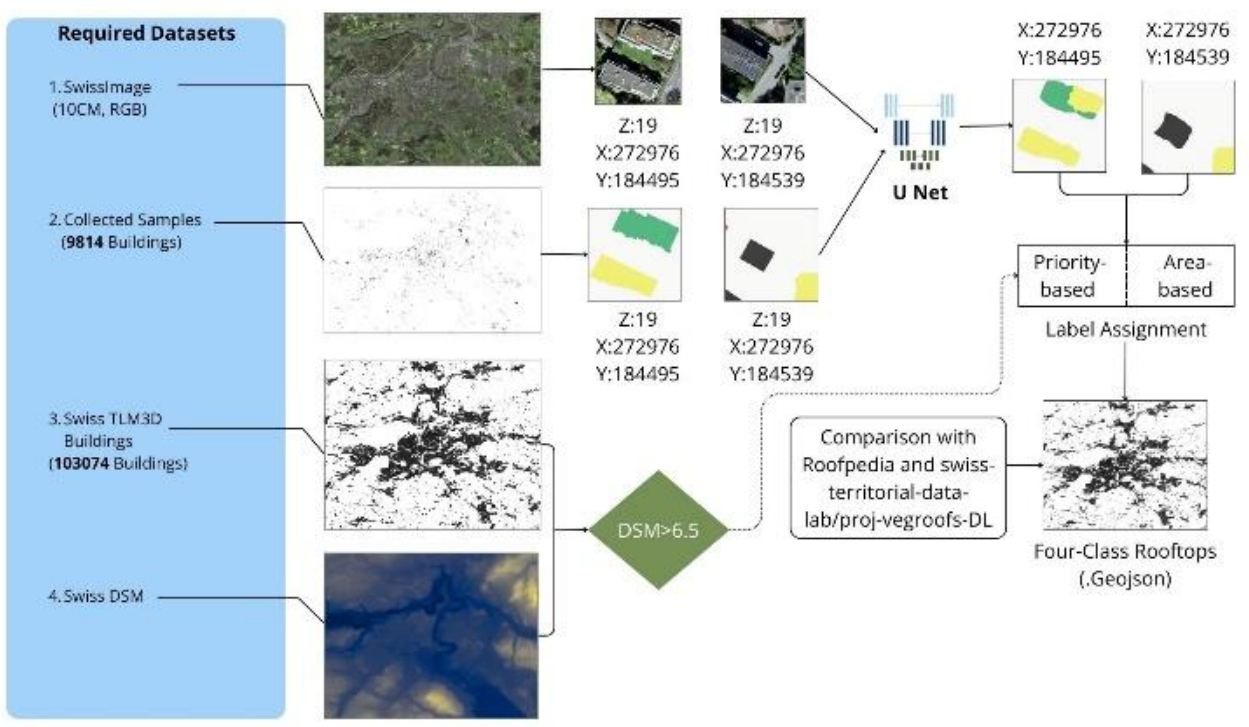

*Fig. 2: Study Workflow of Modified Roofpedia*

**Stage 2: Flat Roof Filtering.** QGIS Zonal Statistics is used to compute the average DSM slope per building. Buildings with average slope below 6.5 degrees are classified as flat and retained for further analysis. This step reduces the full building stock to a flat-roof subset that forms the input for deep learning classification.

**Stage 3: Rooftop Classification.** The flat-roof subset is passed through a modified U-Net semantic segmentation model with a ResNet-50 encoder backbone pretrained on ImageNet. Custom class weights are computed from the training dataset to address class imbalance. The model classifies each rooftop pixel into one of four categories ( Fig. 3): Green (currently vegetated rooftops providing active cooling), Potential Green (flat rooftops suitable for green roof installation), Solar (flat rooftops occupied by photovoltaic panels), and Not Suitable (flat rooftops where installation is not feasible due to rooftop materials or alternative recreational use).

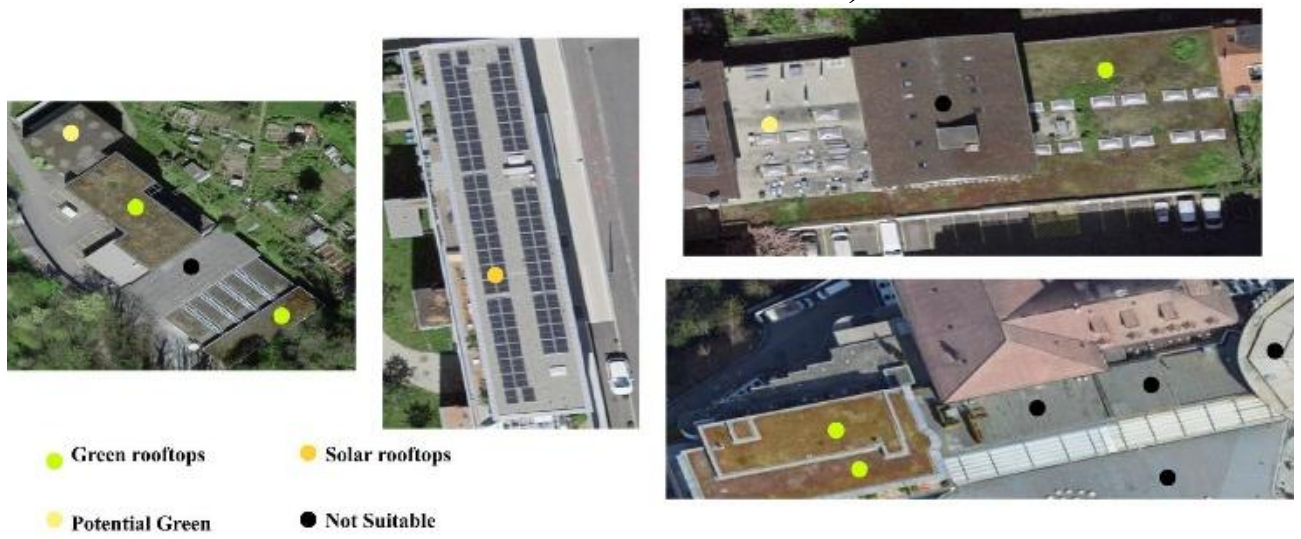

*Fig. 3: SAMPLES of Rooftops in each class*

**Stage 4: Label Assignment.** Predicted patches are spatially intersected with swissTLM3D building footprints. Where multiple classes overlap a single building, one of two label assignment strategies is applied (Fig. 4). The priority-based strategy assigns a single label following a fixed decision hierarchy: Solar takes precedence, followed by Potential Green, then Green, and finally Not Possible. The area-based strategy assigns the class with the largest intersected area proportion within each building boundary. The final output is a four-class rooftop classification dataset stored as a GeoJSON file, suitable for semantic segmentation, green rooftop detection, and ecosystem service assessment.

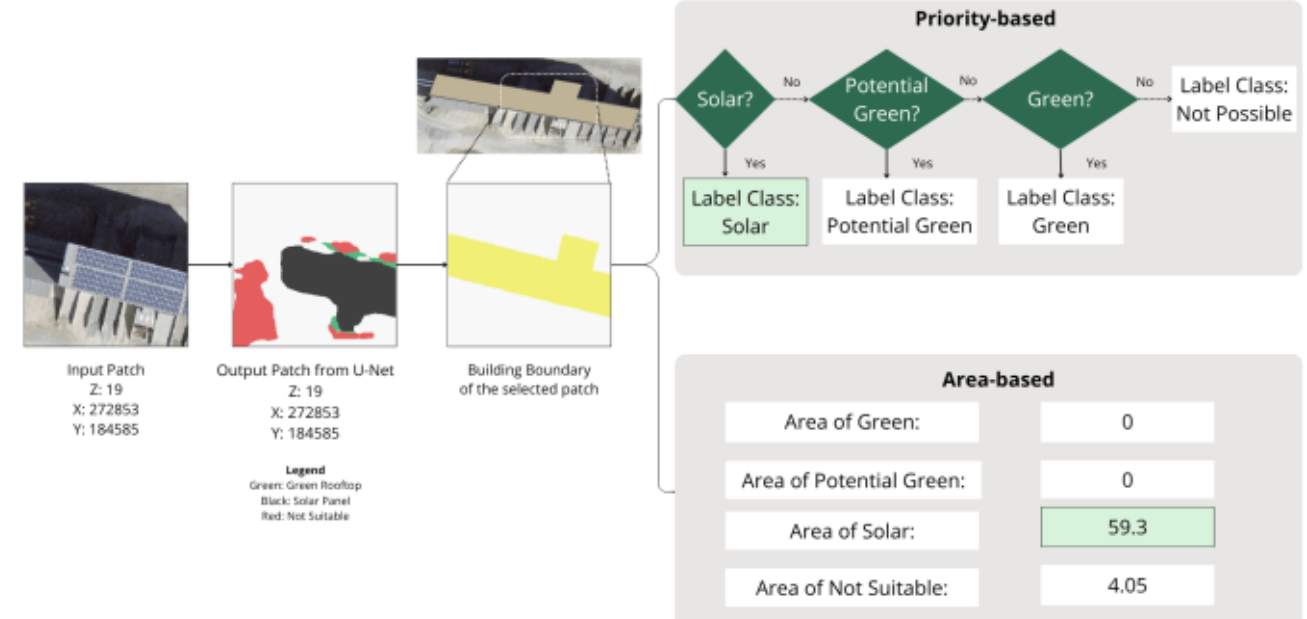

*Fig. 4: Labeling Strategies*

## G. *Experimental Setup*

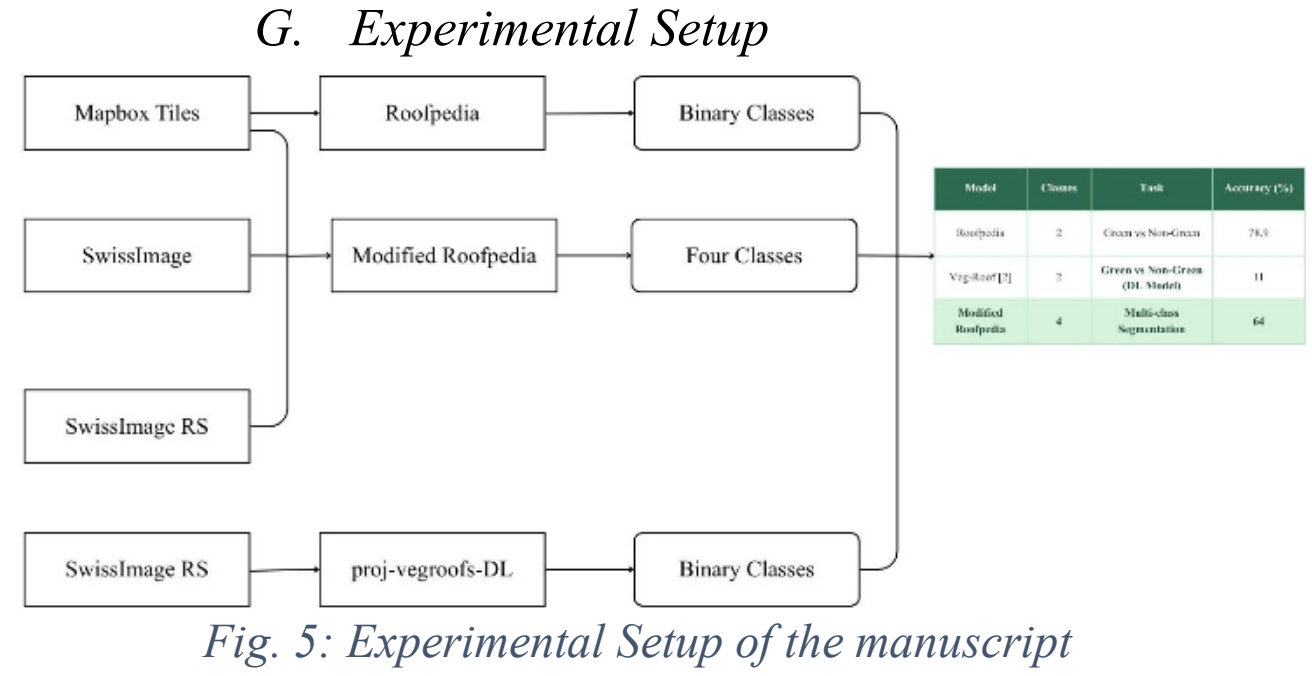

*Fig. 5: Experimental Setup of the manuscript*

Five configurations across three deep learning model families are compared in this study, as illustrated in Fig. 5. The original Roofpedia model uses Mapbox tiles as input and performs binary classification, distinguishing green from non-green rooftops. The proj-vegroofs-DL model uses SwissImage RS as input and performs two-class vegetation segmentation, as described in Section III-C. The proposed Modified Roofpedia model is evaluated with three input datasets (Mapbox, SwissImage, and SwissImage RS) and performs four-class segmentation across the categories defined in Section III-F (3). Together these give five end-to-end pipelines: Roofpedia (Mapbox); Modified Roofpedia (Mapbox, SwissImage, and SwissImage RS); and proj-vegroofs-DL (SwissImage RS). All pipelines are evaluated on the same rooftop-classification task domain, enabling a direct comparison of performance.

# IV. Results and evaluation

This section reports on the experimental results in two parts. Experiment I evaluates the proposed Modified Roofpedia model separately on each of the three aerial image sources (Mapbox, SWISSIMAGE, and SWISSIMAGE RS) to quantify the influence of the input imagery on the four-class segmentation task. Experiment II places all five trained pipelines side by side, namely the original Roofpedia model, the three Modified Roofpedia configurations, and the proj-vegroofs-DL model, to compare their learning behavior, convergence, and generalization under a common training budget of 200 epochs. The section closes with the city scale green roof potential estimation for Bern and an evaluation conclusion based on the distribution of false positives across pipelines. All models were trained with the metrics defined in Section III-D, and unless stated otherwise, the reported values refer to the pixel level segmentation performance on the held-out validation set.

*A. Experiment I: Modified Roofpedia Model*

*1) Modified Roofpedia Model with Mapbox Dataset*

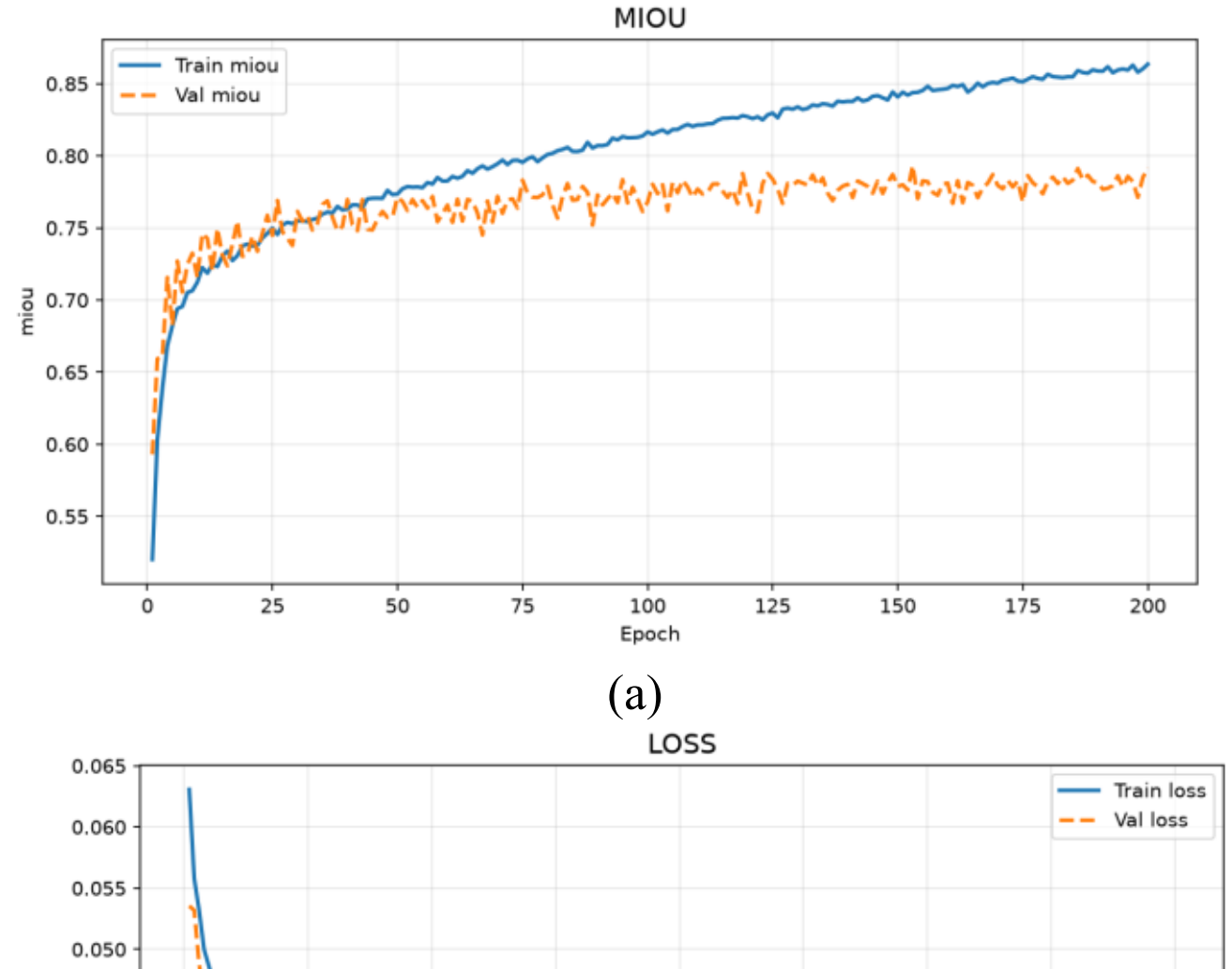

(a)

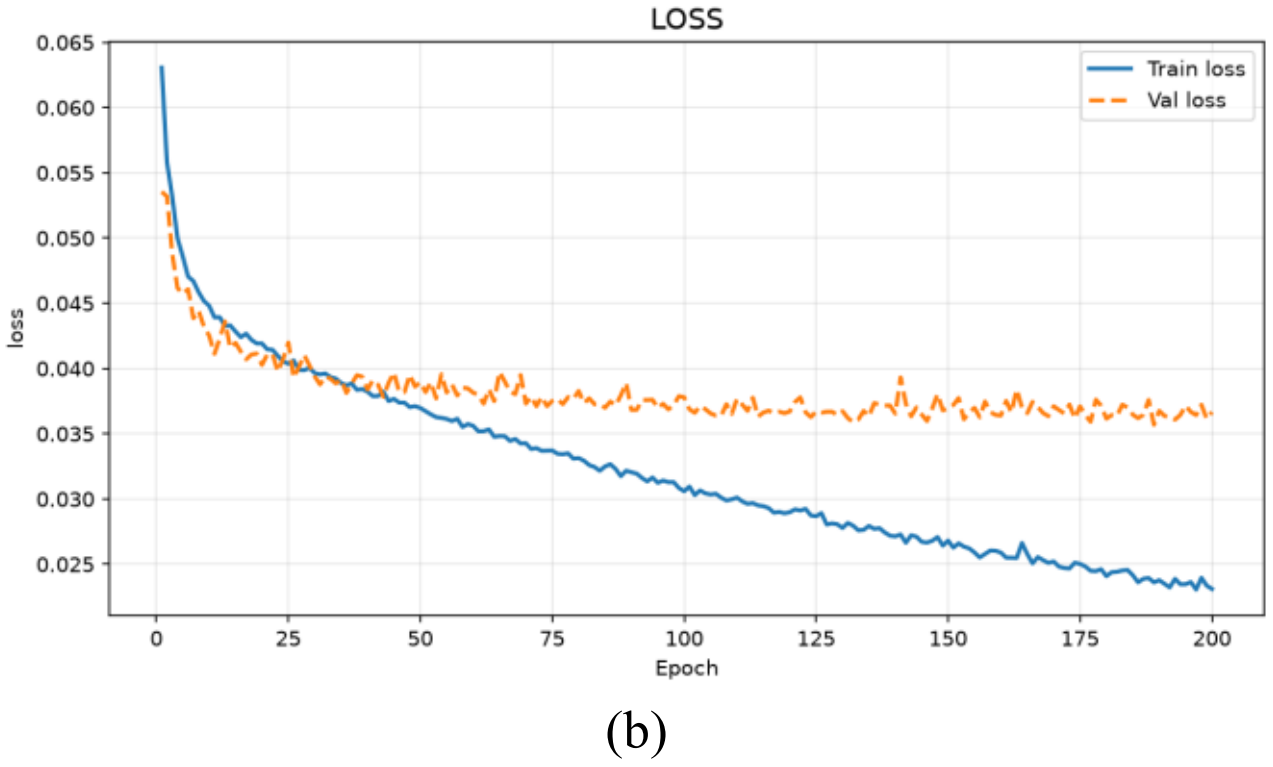

(b)

*Fig 6: Training and Validation Logs of Modified Roofpedia Model with Mapbox dataset*

Fig 6 (a) and (b) present the training and validation mIoU and loss curves for the Modified Roofpedia model trained on Mapbox tiles over 200 epochs. Three phases can be distinguished in the learning behaviour. In the first phase, covering approximately epochs 0 to 25, the training mIoU increases rapidly from approximately 0.53 to 0.75, and the validation mIoU tracks it almost exactly. During this phase the model acquires the dominant, transferable features of the task: the spectral separation of vegetated surfaces from mineral roofing materials, the geometric regularity of photovoltaic arrays, and the characteristic texture of gravel and bitumen membranes. The close agreement between the two curves indicates that these features generalize from the training tiles to unseen validation tiles.

In the second phase, from approximately epoch 25 to 30 onward, the two curves separate. The training mIoU continues to climb at a slower but steady rate, reaching approximately 0.86 by epoch 200, whereas the validation mIoU settles into a narrow band between 0.77 and 0.79 and finishes at approximately 0.78. The third phase is therefore a long plateau in which further optimization benefits only the training set. The loss curves in Fig. 6 mirror this pattern: both losses fall sharply from an initial value of approximately 0.063 to approximately 0.040 within the first 25 epochs, after which the training loss decreases monotonically to approximately 0.024 by epoch 200 while the validation loss stabilizes between 0.036 and 0.039 with no further sustained reduction. The validation loss never exhibits a clear upward trend, which distinguishes this behaviour from severe overfitting; the model does not progressively degrade on unseen data, but the generalization gap of approximately 0.08 mIoU that opens after epoch 30 persists to the end of training. From a practical standpoint, the Mapbox configuration reaches essentially its final validation performance within the first 30 epochs, and the remaining 170 epochs contribute only training set fit.

*2) Modified Roofpedia Model with Swisstopo Dataset*

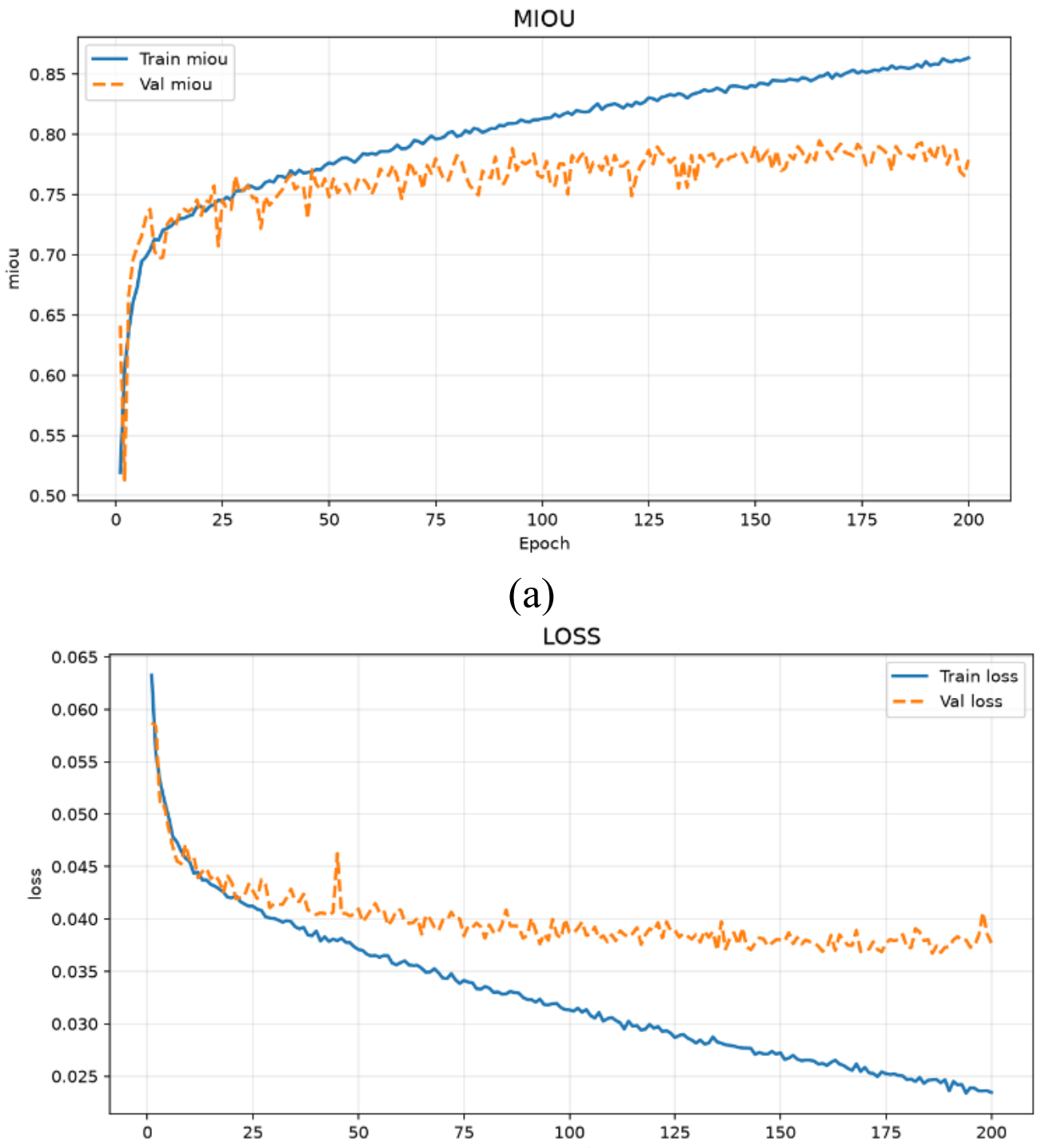

(b)

*Fig 7: Training and Validation Logs of Modified Roofpedia Model with SwissImage dataset*

Fig 7 (a) and (b) present the corresponding curves for the Modified Roofpedia model trained on the SWISSIMAGE dataset. The overall learning behaviour closely resembles the Mapbox configuration despite the substantially higher spatial resolution of the input (10 cm versus approximately 50 cm). The training mIoU rises rapidly from approximately 0.52 to 0.75 within the first 25 to 30 epochs and then continues to improve gradually, reaching approximately 0.86 by epoch 200. The validation mIoU climbs in parallel from approximately 0.51 to between 0.74 and 0.76 by epoch 25 to 30, after which it plateaus in a band between 0.75 and 0.79 and finishes at approximately 0.78.

The loss curves show the same structure. The training loss decreases sharply from approximately 0.063 to 0.040 over the first 25 epochs and then declines steadily and monotonically to approximately 0.023 by epoch 200. The validation loss falls in parallel from approximately 0.058 to 0.043 during the initial epochs and then fluctuates within a band between 0.037 and 0.040 for the remainder of training. A single small transient spike to approximately 0.046 occurs around epoch 45, after which the validation loss returns to its plateau band within a few epochs. This spike coincides with no corresponding disturbance in the training curves and is therefore attributed to a difficult validation batch rather than to an optimization problem. The final validation mIoU of approximately 0.78 is statistically indistinguishable from the Mapbox result, indicating that for the four class formulation the higher spatial resolution of SWISSIMAGE does not, by itself, translate into a higher aggregate segmentation score. As shown in Section IV-D, however, the two configurations distribute their errors across the classes very differently.

*3) Modified Roofpedia Model with SwissImage RS Dataset*

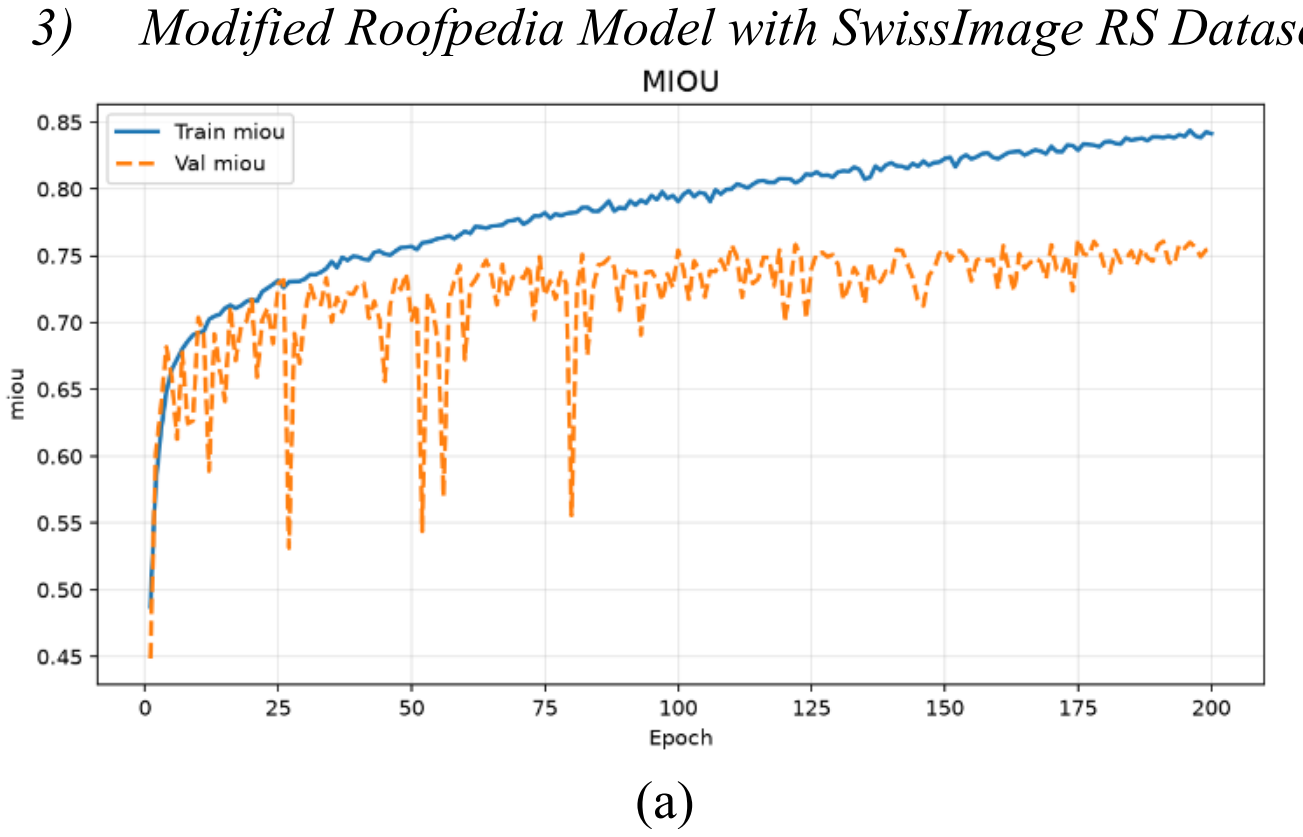

(a)

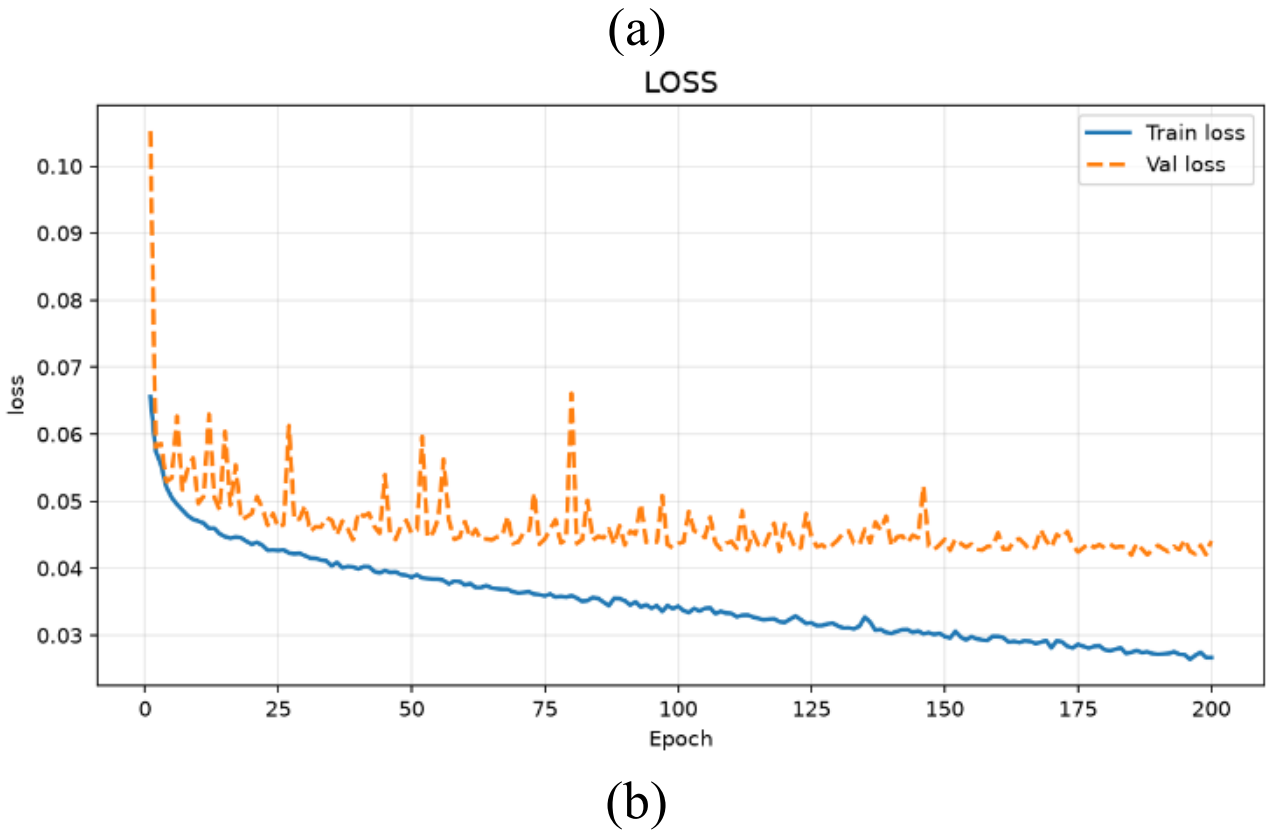

(b)

*Fig. 8: Training and Validation Logs of Modified Roofpedia Model with SwissImage RS dataset*

Fig. 8 (a) and (b) present the training and validation curves for the Modified Roofpedia model trained on the four band SWISSIMAGE RS dataset, which adds a near infrared channel to the red, green, and blue bands. The learning trajectory again follows the now familiar two phase pattern. The training mIoU rises from approximately 0.52 to 0.75 within the first 25 to 30 epochs and reaches approximately 0.86 by epoch 200. The validation mIoU tracks the training curve closely during the initial phase, rising from approximately 0.51 to between 0.74 and 0.76 by epoch 25 to 30, and then plateaus in the band between 0.75 and 0.79, finishing at approximately 0.78. The training loss decreases from approximately 0.063 to 0.040 over the first 25 epochs and continues down to approximately 0.023 by epoch 200, while the validation loss falls from approximately 0.060 to 0.043 and then fluctuates between 0.037 and 0.040, with one transient spike to approximately 0.046 around epoch 45.

The addition of the near infrared band therefore does not change the aggregate validation performance of the four-class task, which remains at approximately 0.78 mIoU across all three input sources. This consistency suggests that the ceiling on validation performance is imposed not by the spectral or spatial richness of the imagery but by the intrinsic ambiguity of the class definitions, in particular the boundary between Potential Green and Not Suitable, which depends on structural roof properties that are not directly observable in any optical imagery. The per class error analysis in Section IV-D supports this interpretation: the near infrared band sharply reduces confusion involving existing vegetation while shifting the residual errors into the Potential Green class.

*4) Conclusion of Experiment I*

Taken together, the three single source configurations show that the Modified Roofpedia architecture converges to the same validation performance of approximately 0.78 mIoU and 0.036 to 0.040 validation loss regardless of whether the input is commercial Mapbox imagery at approximately 0.5 m resolution, national SWISSIMAGE orthophotos at 0.1 m resolution, or four band SWISSIMAGE RS data. In all three cases the model reaches its final generalization level within the first 25 to 30 epochs, and training for the full 200 epochs widens the train validation gap to approximately 0.08 mIoU without improving predictive performance. Early stopping around epoch 30 to 50, or stronger regularization through increased dropout, weight decay, or data augmentation, would therefore preserve validation performance while reducing the training cost by more than 80%.

### *B. Experiment II: Comparison of different deep learning Rooftop mapping pipelines*

Fig. 9 (a) to (d) present the training mIoU, training loss, validation mIoU, and validation loss of all five pipelines: the original Roofpedia model, the three Modified Roofpedia configurations from Experiment I, and the proj-vegroofs-DL model. Three groups are clearly separated into all four panels. The original Roofpedia model occupies the top of the validation mIoU panel, the three Modified Roofpedia configurations form a tight middle band, and proj-vegroofs-DL runs distinctly below the others in both mIoU panels and above them in both loss panels. This ordering follows the number of classes each pipeline must separate: two for Roofpedia, four for Modified Roofpedia, and two for proj-vegroofs-DL. The following subsections describe each pipeline in detail.

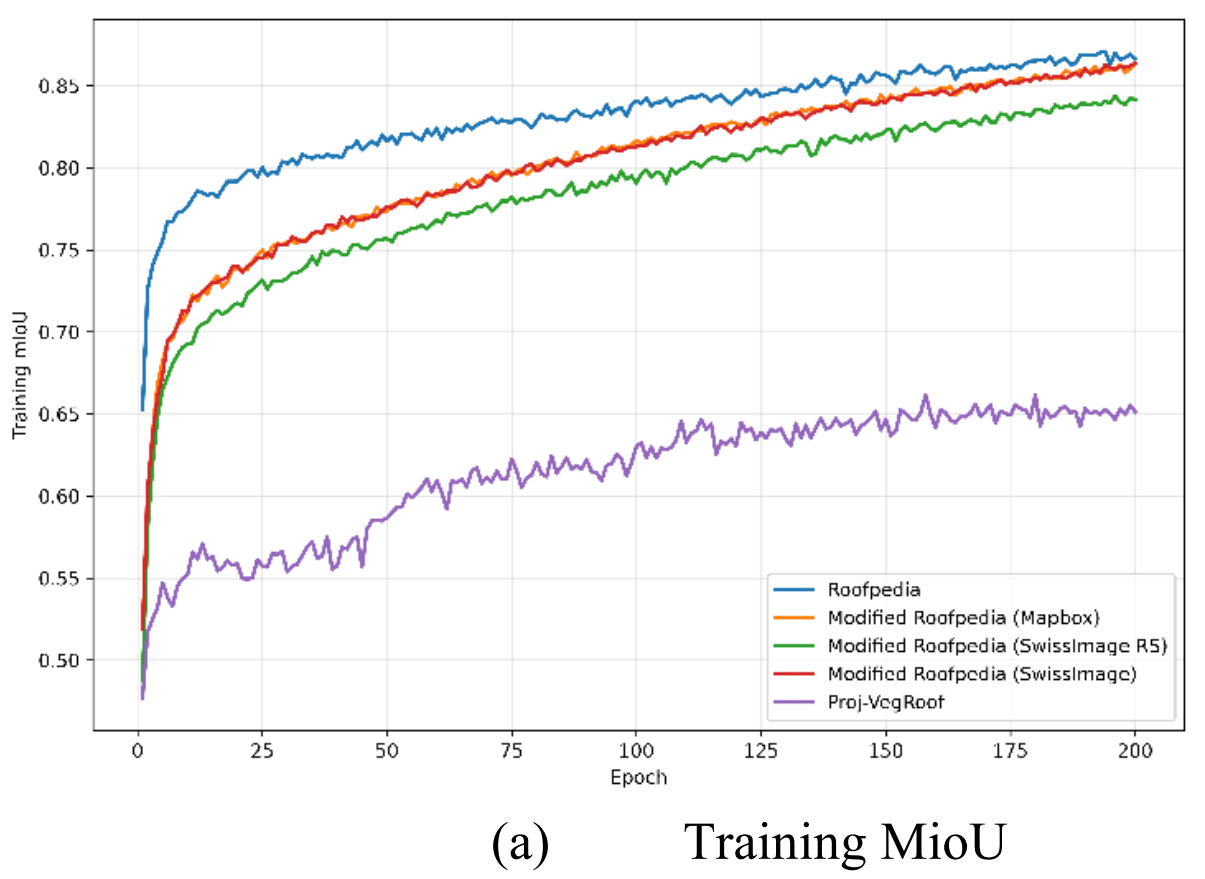

(a) Training MioU

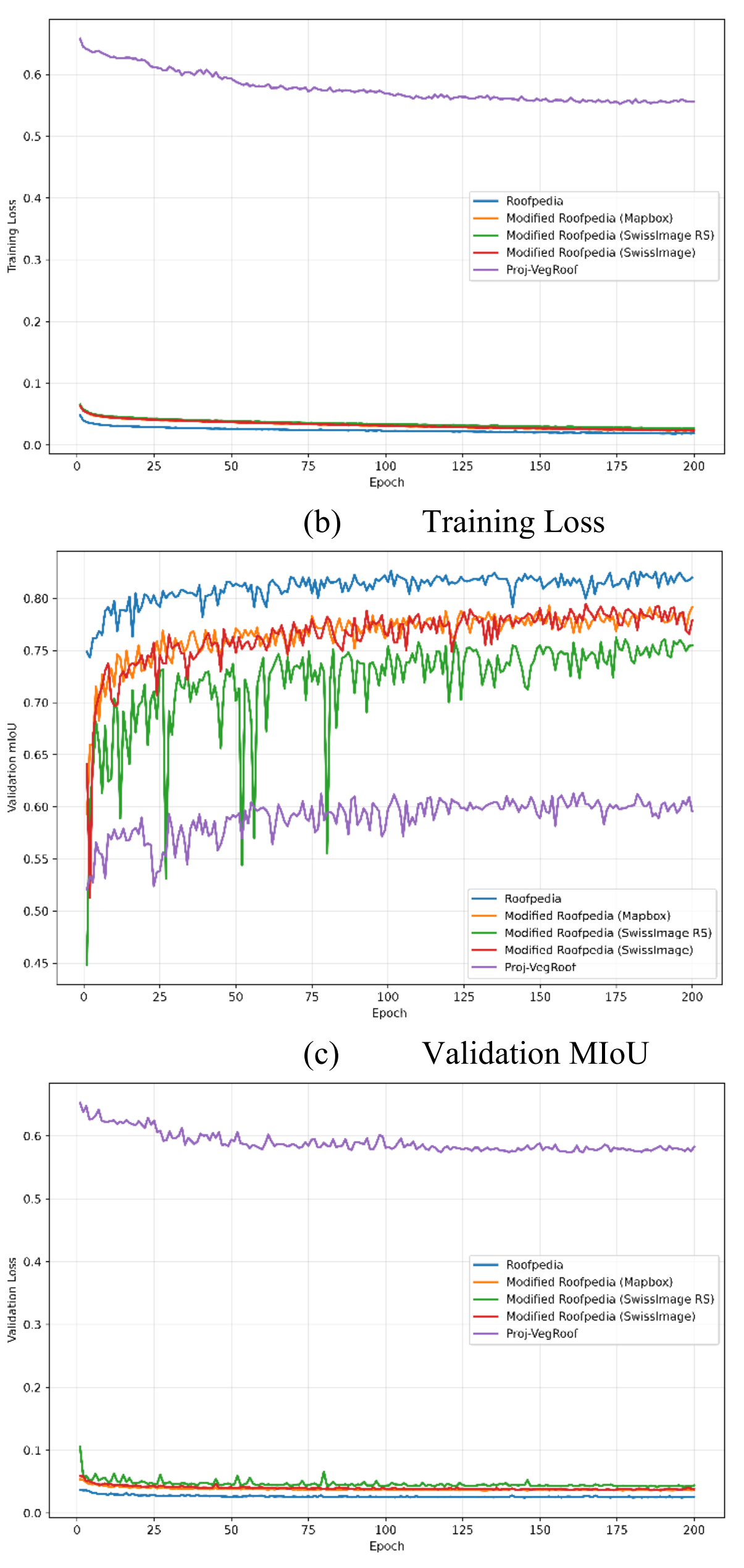

(b) Training Loss

(c) Validation MIoU

(d) Validation Loss

*Fig. 9: (a) Training mIoU (b) Training loss (c) Validation MIoU and (d) Validation Loss of five deep learning pipelines.*

TABLE 2 summarises the comparison. The original Roofpedia model achieves 78.9% accuracy on a binary task. The proj-vegroofs-DL model achieves 11% accuracy, also on a binary task. 'Bare' class of proj-vegroofs-DL model is mapped as 'Potential Green' and 'Vegetation' class is mapped as 'Green' here. There is the risk of involving Potential Green and 'Not suitable' roofs in 'Bare' class of proj-vegroofs-DL output. The proposed Modified Roofpedia model achieves 64% accuracy while performing the substantially harder four-class task. Because the baselines and the proposed model solve tasks of different difficulty, these values are not directly comparable; the four-class model's value lies in producing all four policy-relevant categories from a single pass, as the building-level analysis in Section IV-C quantifies.

*TABLE 2*
*SUMMARY OF TRAINING RESULTS*

| Model | Classes | Task | Accuracy (%) |
|---|---|---|---|
| Roofpedia [4] | 2 | Green vs Non-Green | 78.9 |
| Modified Roofpedia (Mapbox) | **4** | **Four-class Segmentation** | **64** |
| Modified Roofpedia (SwissImage) | | | 60 |
| Modified Roofpedia (SwissImage RS) | | | 37 |
| proj-vegroofs-DL [3] | 2 | Bare vs Vegetation (DL Model) | 11 |

*Note: Accuracy values in this table are pixel-level overall accuracy on the held-out validation split (Section III-D). They are not directly comparable to the building-level accuracies reported in Section IV-C to E (Tables 4 to 6), which are computed against the full reference-labelled building set after Stage 4 label assignment. The proj-vegroofs-DL value is low because its binary output maps poorly onto the two-class reference used here.*

*1) Roofpedia*

As presented in Fig. 9 (a) to (d), the training and validation mIoU and loss curves for the original Roofpedia model can be seen. The training mIoU rises rapidly from approximately 0.65 to 0.80 within the first 25 epochs and then increases gradually and steadily, reaching approximately 0.87 by epoch 200. The validation mIoU closely tracks the training curve during the initial phase, rising from approximately 0.745 to 0.80 by epoch 25 to 30. Beyond this point it plateaus, fluctuating within a relatively narrow band between 0.80 and 0.82 with a few minor transient dips, for example around epochs 140 and 165, and finishes at approximately 0.82.

The loss curves show the same structure. The training loss decreases sharply from approximately 0.048 to 0.030 over the first 25 epochs and then declines steadily and monotonically to approximately 0.019 by epoch 200. The validation loss falls in parallel from approximately 0.036 to 0.027 before plateauing between 0.025 and 0.028, with small transient spikes around epochs 80, 140, and 165. Roofpedia thus achieves the highest absolute validation scores of all five pipelines, but this must be read against its task: distinguishing green from non green rooftops is a considerably easier problem than the four formulations. The high starting point of both curves, with a training mIoU of 0.65 already at epoch 0, also reflects the lower complexity of the binary decision boundary. The gap between the final training mIoU of 0.87 and validation mIoU of 0.82 shows the same moderate overfitting pattern observed in the Modified Roofpedia experiments, though the gap of 0.05 is the smallest of the five pipelines.

*2) proj-vegroofs-DL*

Fig. 10 (a) to (d) present the training and evaluation IoU and loss curves for the proj-vegroofs-DL model over 200 epochs. The training IoU increases steadily from an initial value of approximately 0.50, reaching approximately 0.56 to 0.57 by epoch 25. A brief plateau follows, after which the training IoU shows a marked increase around epochs 45 to 50, rising to approximately 0.60, and subsequently continues to improve steadily for the remainder of training, reaching approximately 0.66 to 0.67 by epoch 200. The evaluation IoU tracks the training curve closely during the first 40 epochs but exhibits a sharp transient drop to approximately 0.34 at around epoch 45. Following this event the evaluation IoU recovers and plateaus, fluctuating between 0.55 and 0.58 for the remainder of training and ending at approximately 0.58.

The loss curves display the corresponding pattern. Both training and evaluation loss decrease from an initial value of approximately 0.65 during the first 25 epochs. Coinciding with the IoU anomaly, the evaluation loss spikes to approximately 0.755 at around epoch 45 before recovering and stabilizing between 0.60 and 0.62. The training loss decreases steadily throughout, reaching approximately 0.55 by epoch 200 without exhibiting the transient instability of the evaluation curve. The transient drop at epoch 45 is attributed to a learning rate warm restart in the cosine annealing scheduler, after which the model recovers within one epoch and continues to improve stably. Beyond this event, the persistent and widening gap between training and evaluation performance reflects the difficulty of the bare-versus-vegetation distinction on this dataset: the model must separate vegetated from bare rooftops (the latter standing in for potential green roofs here) from relatively few training samples, which makes it more prone to memorizing training-specific patterns. Evaluation performance saturates early, at around epoch 50, and shows limited further gains, so early stopping combined with stronger regularization or greater training data diversity would improve efficiency at no cost in accuracy.

*3) Performance of Modified Roofpedia on Mapbox, SWISS IMAGE and SWISS IMAGE RS*

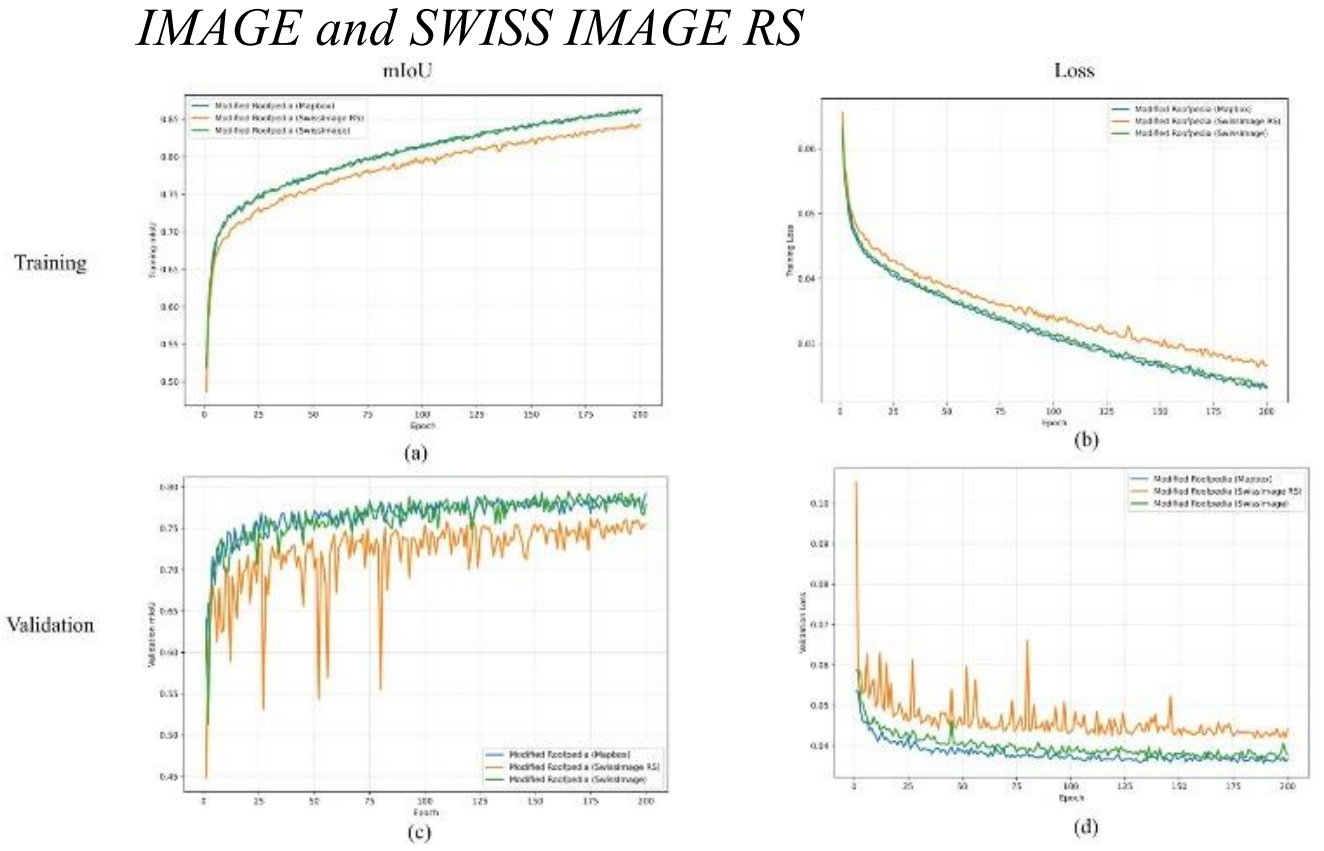

*Fig. 10: (a) Training mIoU (b) Training loss (c) Validation MioU and (d) Validation Loss of Modified Roofpedia pipelines with different source of images*

Across all three single source configurations (Fig. 10), the close alignment between training and validation curves during the early epochs (0 to 25) indicates that the model learns generalizable features during this phase, regardless of the input imagery. Beyond epoch 25 to 30, however, the curves progressively diverge: training mIoU and loss continue to improve while the validation curves plateau, a pattern indicative of mild overfitting. In each configuration the model continues to fit the training data more closely, with training mIoU rising to approximately 0.86 and training loss falling to between 0.023 and 0.024, but this improvement does not translate into better generalization on unseen validation data, which stabilizes at an mIoU of approximately 0.78 and a loss of approximately 0.037 to 0.040 for Mapbox, SWISSIMAGE, and SWISSIMAGE RS alike.

This behaviour suggests that, beyond approximately epoch 25 to 30, additional training primarily improves model fit to the training set rather than genuine predictive performance. The relatively stable and narrow fluctuation band of the validation curves, rather than a clear upward trend in validation loss, indicates that the overfitting is moderate rather than severe. The model does not catastrophically memorize the training data, but the generalization gap persists across the remaining 170 epochs of training. Early stopping around epoch 25 to 30, or the application of stronger regularization such as increased dropout, weight decay, or data augmentation, could therefore achieve comparable validation performance with substantially reduced training time.

*C. Results of green roof potential estimation*

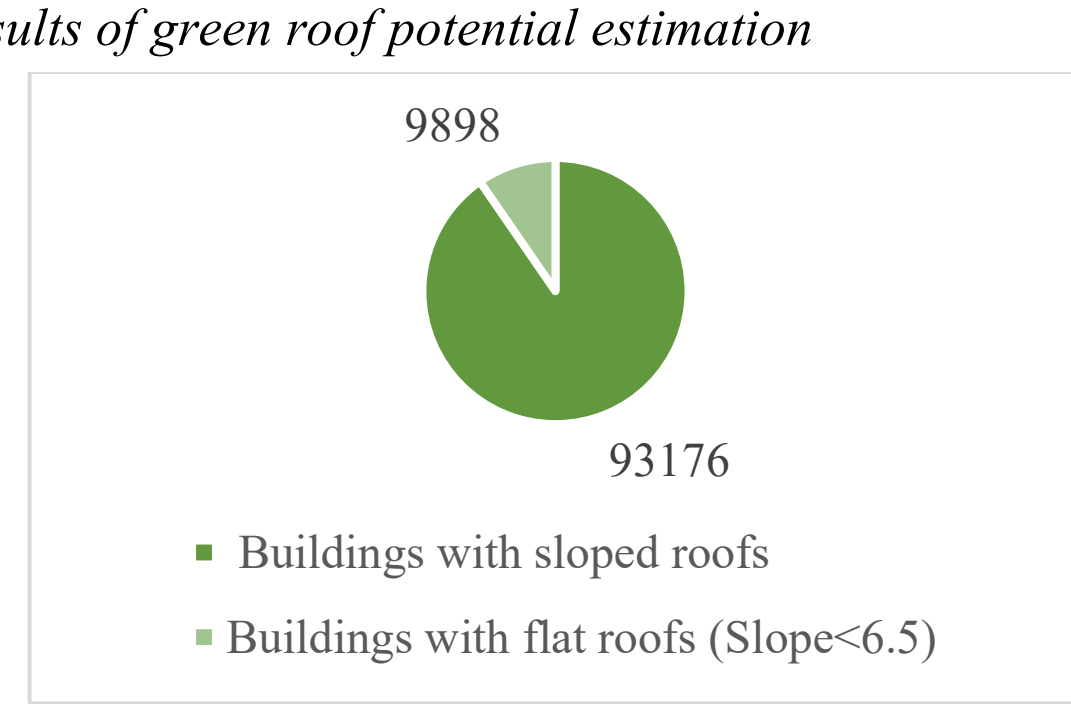

*Fig. 11. Building Stock Analysis*

The best performing configuration was applied to the full flat roof building stock of Bern to produce the city scale rooftop classification. Fig. 11 (Building Stock Analysis) shows the outcome of the slope-based filtering stage: of the 103,074 building footprints in the study area, the 9,898 buildings (approximately 9.6% of the stock) whose average surface slope computed from the swissSURFACE3D model is below 6.5º were retained as the flat-roof subset; the remaining 93,176 were classified as sloped and excluded from further analysis.

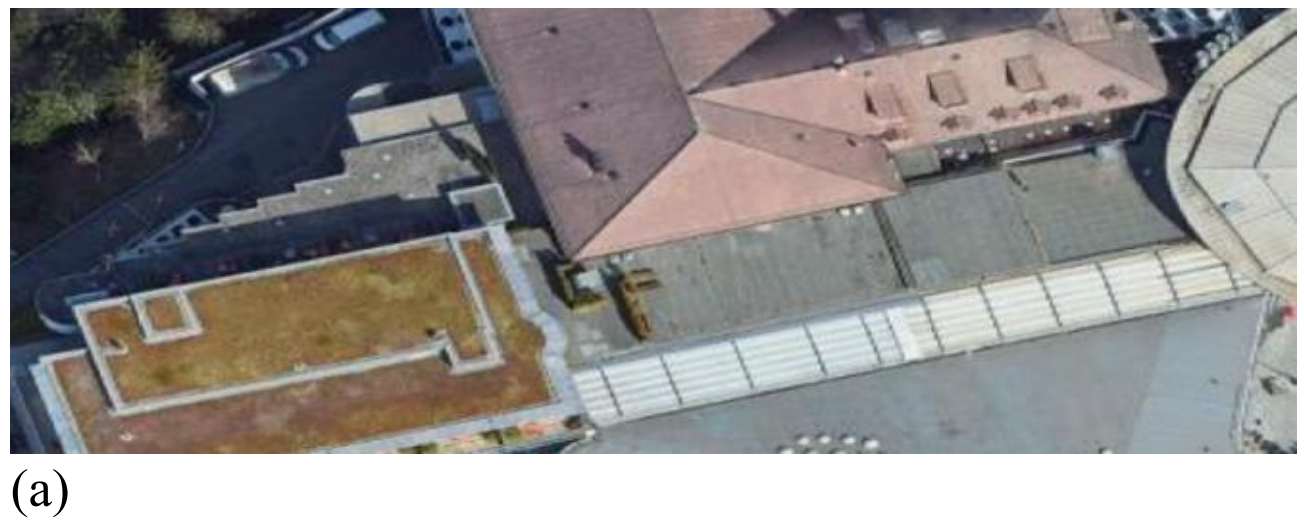
(a)

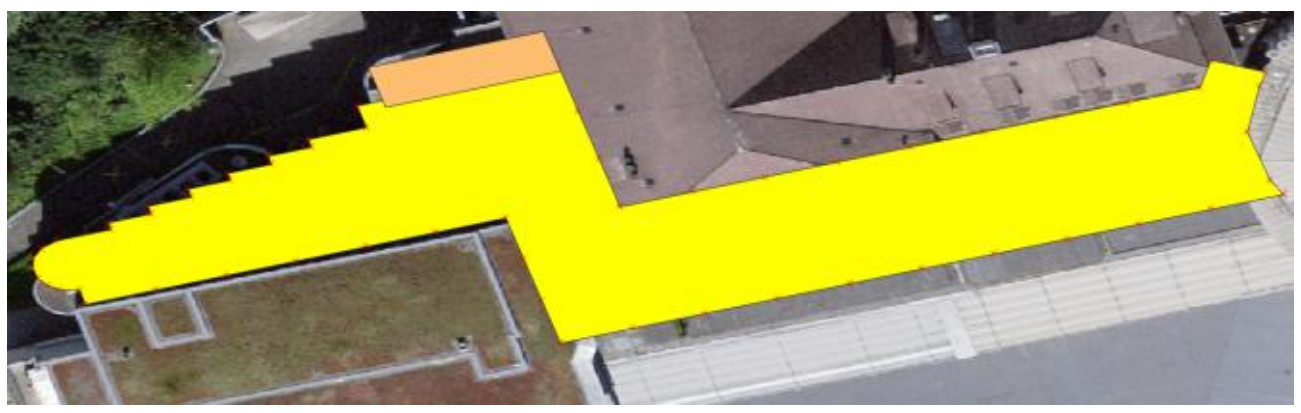
(b)

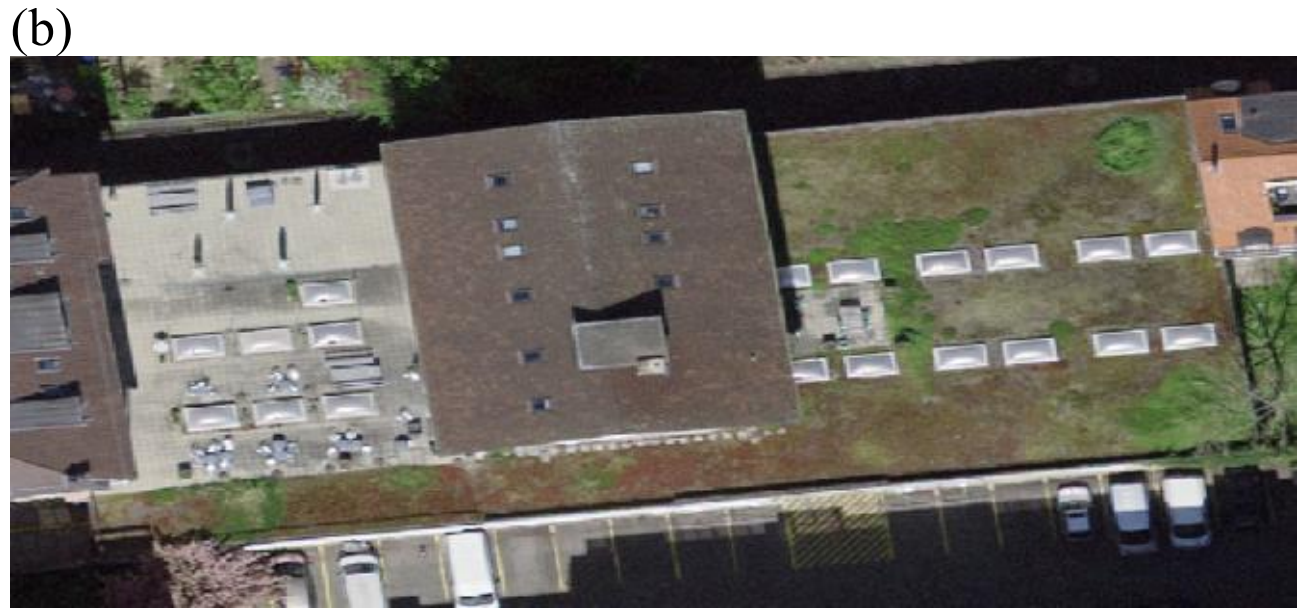
(c)

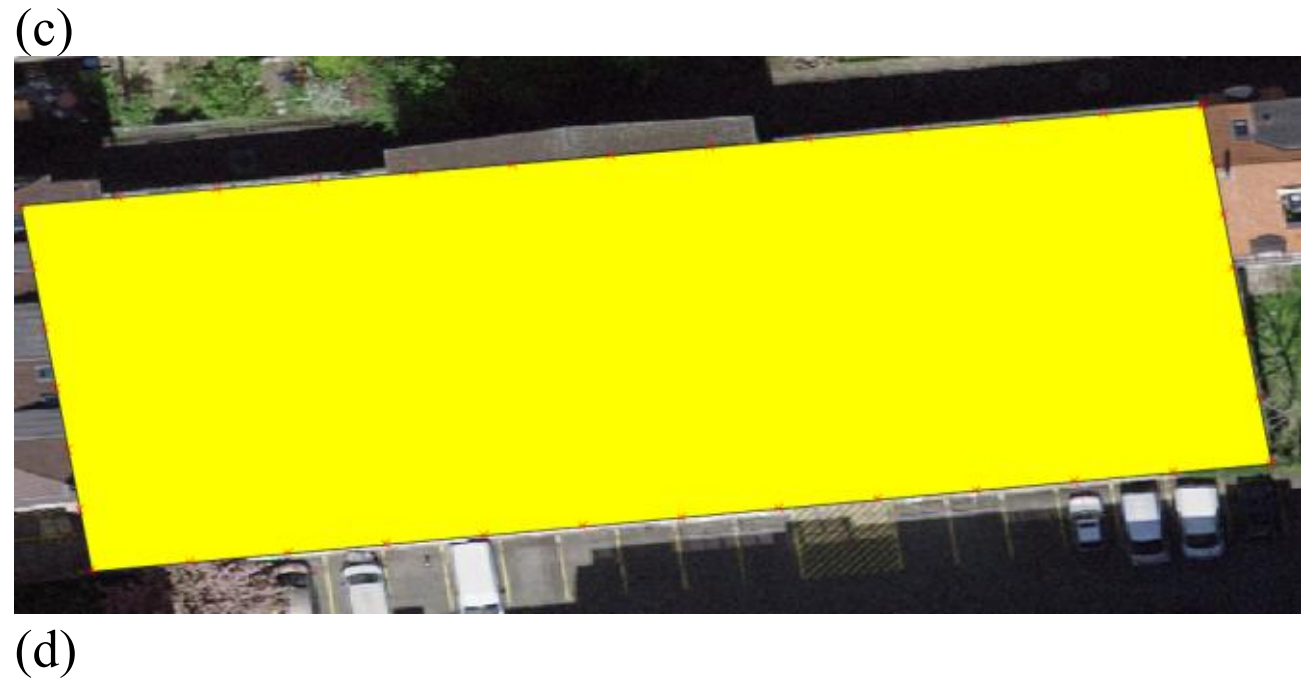
(d)

*Fig. 12: A single building footprint can span several predicted roof classes. Panels (a) and (c) show the aerial imagery; panels (b) and (d) show the corresponding swissTLM3D building polygon, within which more than one class co-occurs. The area-based labelling strategy resolves this by assigning each footprint the class with the largest intersected area.*

Fig. 13 compares the class distributions obtained with the two label assignment strategies applied to the same pixel predictions. Under the priority-based strategy, in which Solar takes precedence over Potential Green, Green, and Not Suitable, the Solar class receives 11.7 % of buildings and the Potential Green class 44.2 %. Under the area-based strategy, in which each building receives the class with the largest intersected area within its footprint, the Solar share falls to 6.3% and the shares of Green and Not Suitable rise correspondingly. The difference illustrates that the labelling strategy alone, applied to identical model outputs, can nearly double the estimated number of solar rooftops, which has direct consequences for policy-oriented reporting.

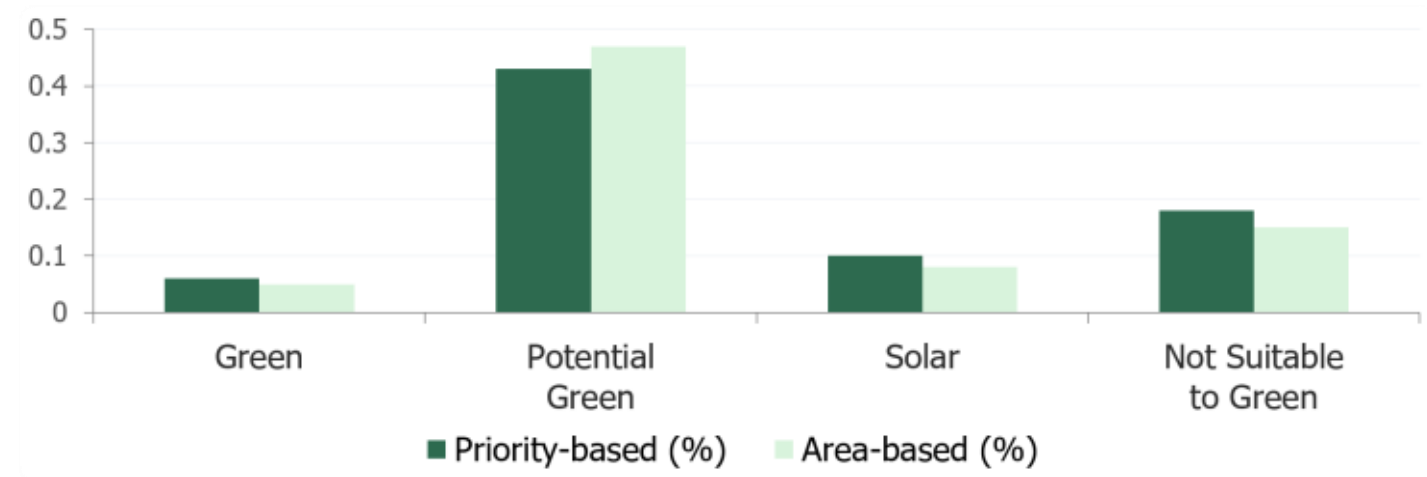

Fig. 13. Class Distribution Comparison of two labelling strategies for the output of Modified Roofpedia with SwissImage

TABLE 3 presents sample training and predicted patches at the location X: 272866, Y: 184526 for the Roofpedia and Modified Roofpedia configurations; the proj-vegroofs-DL pipeline structure does not expose training and predicted sample patches. Fig. 14 shows sample segmentation results over a residential block for all three pipelines (five models), with existing green rooftops in green, potential green rooftops in yellow, solar rooftops in orange, and rooftops not suitable for greening in red; Roofpedia and proj-vegroofs-DL appear as binary segmentations. Visual inspection confirms the quantitative findings: the Modified Roofpedia outputs on Mapbox and SWISSIMAGE delineate building boundaries sharply and assign coherent single class labels to most rooftops, whereas the SWISSIMAGE RS output shows blurred boundaries, and the binary pipelines cannot express the distinction between existing and potential greening at all.

*TABLE 3.*

*Sample of Training and Predicted Results (X: 272866, Y:184526). Please note that the Proj--vegroofs-DL pipeline structure do not provide the training and predicted sample patches.*

| **Model** | **Input Image** | **Correct Label** | **Predicted Label** |
|---|---|---|---|
| Roofpedia | | | |
| Modified Roofpedia (SwissImag e) | | | |
| Modified Roofpedia (Mapbox) | | | |

| Modified Roofpedia (SwissImage RS) | 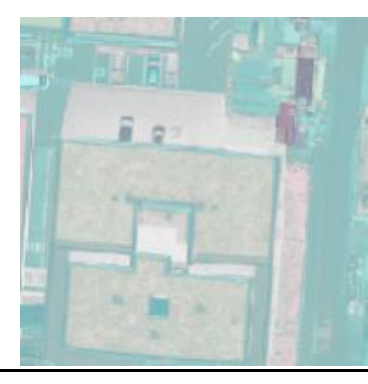 | 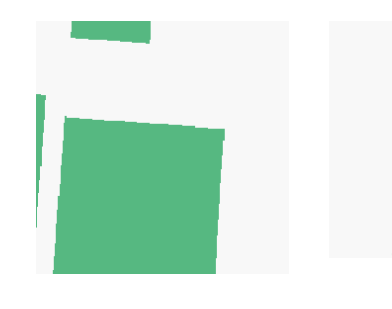 |
|---|---|---|

*D. Result Evaluation*

Across all experiments, the Modified Roofpedia configurations converge to a consistent validation mIoU of approximately 0.78 regardless of the single input source used (SWISSIMAGE, SWISSIMAGE RS, or Mapbox).

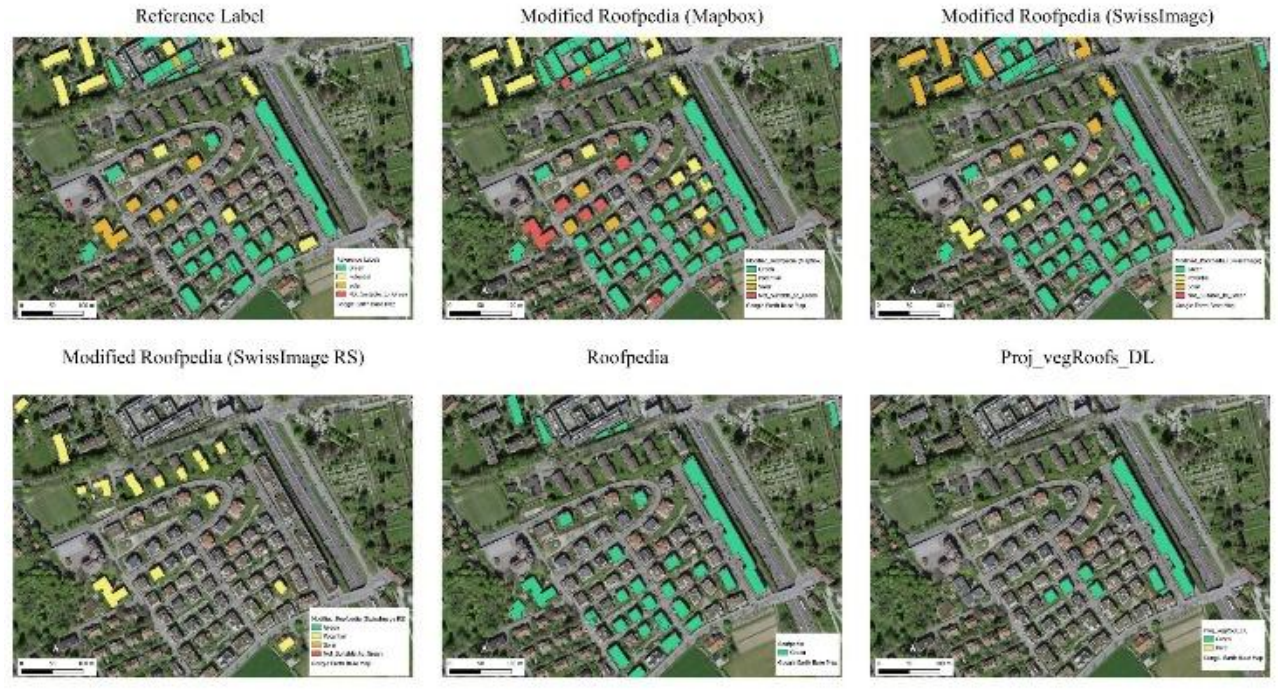

*Fig. 14: Sample segmentation results by using Modified Roofpedia (Mapbox, SwissImage, SwissImage RS), Roofpedia and proj_vegroofs_DL pipelines, respectively (green: existing rooftops, yellow:potential green rooftops, orange: solar and red: not suitable to install green rooftops). Please note that Roofpedia and proj_vegroofs_DL pipelines are for binary class segmentation.*

*TABLE 4.*
*Results per class of each pipelines (Please note that the study region of proj-vegroofs-DL is slightly smaller than the other regions based on the provided images from STDL Team.)*

| **Reference class** | **Roofpedia** | **Mod. Roofpedia (SwissImage)** | **Mod. Roofpedia (Mapbox)** | **Mod. Roofpedia (SwissImage RS)** | **proj-vegroofs-DL** |
|---|---|---|---|---|---|
| Green | 2922 | 3484 | 3307 | 103 | 572 |
| Potential Green | 0 | 13 | 2532 | 821 | 120 |
| Solar | 0 | 13 | 47 | 9 | 0 |
| Not Suitable | 0 | 1539 | 22 | 567 | 0 |
| Total | 2922 | 5049 | 5908 | 1500 | 692 |

*Note: Accuracy values in this table refer to building-level segmentation performance on the held-out validation split used during training (Section III-D), and are not directly comparable to the pixel-level classification accuracies reported in Section IV-C-E (TABLE 2).*

TABLE 4 reports building-level correct classifications against the reference set of 9,898 labelled buildings. Among the four-class configurations, Mapbox achieves the highest overall building-level accuracy (5,908/9898, 59.7%), followed by SwissImage (5,049/9,898, 51%); SwissImage RS trails substantially (1,500/9,898, 15.2%). Roofpedia, constrained to a single class, correctly classifies 2922 of the 3,772 buildings labelled Green in the reference set (77.5% accuracy on its one class) but by construction contributes nothing to the other three categories. proj-vegroofs-DL correctly classifies 692 buildings overall, concentrated in Green (572) given its vegetation-typology training objective.

Notably, Mapbox and SwissImage diverge sharply in *where* their correct detections concentrate: SwissImage recovers far more Not Suitable buildings correctly (1,539 vs. 22 for Mapbox), while the two are more comparable on Green and Solar. This suggests the two RGB sources are not interchangeable for the four-class task even though their aggregate accuracy is similar.

*E. Misclassification of Modified Roofpedia Models*

TABLE 5 summarizes the false positives of each pipeline, that is, the buildings assigned a positive class that the four-class reference set assigns to a different class. TABLE 6 reports the complementary missed detections, the buildings belonging to a class that the pipeline failed to recover. Reading the two tables together is essential, because a low false positive count can reflect either strong discrimination or a tendency to under-predict, and only the missed-detection table distinguishes the two.

The original Roofpedia model produced the fewest false positives (1,268), concentrated in the Potential Green class (63.2%). This reflects its binary formulation, which cannot separate existing greening from greening potential. Its missed-detection total, however, is high (5,708), spread across Potential Green (42.4%) and Not Suitable (31.9%), confirming that its low false positive count stems from a restricted label space rather than from superior discrimination.

The Modified Roofpedia model trained on SwissImage produced 4,406 false positives, dominated by the Potential Green class (69.1%), indicating that separating rooftops that are already green from those that could be greened is the model's hardest distinction even with 10 cm RGB imagery. Crucially, this configuration achieved by far the lowest missed-detection total of all pipelines (443), meaning that its errors are overwhelmingly confusions between the two green-related classes rather than failures to detect rooftops at all. This combination of near-complete recall with errors concentrated in a single, conceptually adjacent class marks SwissImage as the most balanced configuration.

The Mapbox configuration produced 3,711 false positives, dominated by the Not Suitable class (51.6%). At approximately 50 cm resolution, roofing materials with vegetation-like texture, such as gravel and weathered bitumen, are frequently mistaken for greenable surfaces. Its missed-detection total is also low (279), the second lowest overall, so like SwissImage it detects rooftops reliably and errs mainly in class assignment.

The SwissImage RS configuration produced the fewest false positives among the four-class configurations (2,019), but this must be read alongside its very high missed-detection total (6,379), the second highest of all pipelines. Of its missed detections, 42.0% fall in the Green class and 34.0% in Potential Green, while 49.1% of its false positives fall in Green. The four-band model therefore does not discriminate better than the RGB configurations; instead, it under-predicts, failing to recover a large share of rooftops. Two factors plausibly contribute. First, the near-infrared band strongly separates vegetated from non-vegetated surfaces but adds little information for distinguishing solar panels or roofing materials, so the model leans on the vegetation signal at the expense of the other class boundaries. Second, as noted in the qualitative assessment, the SwissImage RS imagery is visibly blurrier at building scale, which degrades the sharp geometric cues on which solar detection in particular depends. The near-infrared channel is therefore an asset for vegetation-oriented tasks but, used alone at this image quality, a liability for the balanced four-class formulation.

The proj-vegroofs-DL model produced 2,070 false positives, spread mainly across Not Suitable (39.4%) and Potential Green (26.3%), together with the highest missed-detection total of any pipeline (7,136). The STDL developed this model for two-class and six-class vegetation typologies rather than for the present four-class task, and its binary (bare vs vegetation) variant was selected for this comparative study; it therefore recovers only a fragment of the required classes when its output is mapped onto the four-class reference scheme. This reinforces the central argument of the study: existing pipelines each cover part of the information that urban greening policy requires, whereas the proposed four-class formulation, particularly with SwissImage or Mapbox input, is the only configuration that delivers usable detections across all four categories simultaneously.

Across all pipelines, the Solar class is the most stable error category, contributing between 8.8 and 21.4% of false positives and between 6.8 and 10.8% of missed detections, consistent with the distinctive geometric signature of photovoltaic arrays. Taken together, TABLE 5 and 6 indicate that the SwissImage configuration offers the best overall balance, pairing the lowest missed-detection total with false positives concentrated in a single conceptually adjacent class, while the Mapbox configuration performs comparably but shifts its residual errors toward the Not Suitable class.

*TABLE 5.*
*False positives per reference class and pipeline (building counts and share of each pipeline's total errors). The comparison is based on four-class reference data.*

| **Reference class** | **Roofpedia** | **Mod. Roofpedia (SwissImage)** | **Mod. Roofpedia (Mapbox)** | **Mod. Roofpedia (SwissImage RS)** | **proj-vegroofs-DL** |
|---|---|---|---|---|---|
| Green | 0 (0.0%) | 243 (5.5%) | 417 (11.2%) | 991 (49.1%) | 513 (24.8%) |
| Potential Green | 802 (63.2%) | 3,045 (69.1%) | 586 (15.8%) | 231 (11.4%) | 544 (26.3%) |
| Solar | 249 (19.6%) | 824 (18.7%) | 793 (21.4%) | 280 (13.9%) | 197 (9.5%) |
| Not Suitable | 217 (17.1%) | 294 (6.7%) | 1,915 (51.6%) | 517 (25.6%) | 816 (39.4%) |
| Total | 1,268 | 4,406 | 3,711 | 2,019 | 2,070 |

*TABLE 6.*
*Missed detection per reference class and pipeline (building counts and share of each pipeline's total errors). The comparison is based on four-class reference data.*

| **Reference class** | **Roofpedia** | **Mod. Roofpedia (SwissImage)** | **Mod. Roofpedia (Mapbox)** | **Mod. Roofpedia (SwissImage RS)** | **proj-vegroofs-DL** |
|---|---|---|---|---|---|
| Green | 850 (14.9%) | 45 (10.2%) | 48 (17.2%) | 2,678 (42.0%) | 2,687 (37.7%) |
| Potential Green | 2,419 (42.4%) | 163 (36.8%) | 103 (36.9%) | 2,169 (34.0%) | 2,557 (35.8%) |
| Solar | 618 (10.8%) | 30 (6.8%) | 27 (9.7%) | 578 (9.1%) | 670 (9.4%) |
| Not Suitable | 1,821 (31.9%) | 205 (46.3%) | 101 (36.2%) | 954 (15.0%) | 1,222 (17.1%) |
| Total | 5,708 | 443 | 279 | 6,379 | 7,136 |

TABLE 7 presents representative misclassified samples together with the underlying causes, which fall into three recurring categories. The first category is spectral ambiguity of the roof surface itself. In sample (a), a bare roof covered with moss is misclassified as Green, since the spontaneous moss cover produces a vegetation signal that is spectrally indistinguishable from a green roof. This confusion is conceptually difficult to resolve from optical imagery alone, because the boundary between an unmaintained mossy surface and an extensive green roof is partly a matter of definition rather than of appearance; auxiliary information such as substrate depth or maintenance records would be required to separate the two reliably.

*TABLE 7.*
*Misclassified samples and the explanation of causing the misclassification*

| **Misclassified Samples** | **Reason** |
|---|---|
| | The polygon is misclassified as Green Roof, but it is the bare roof covered with moss. |
| 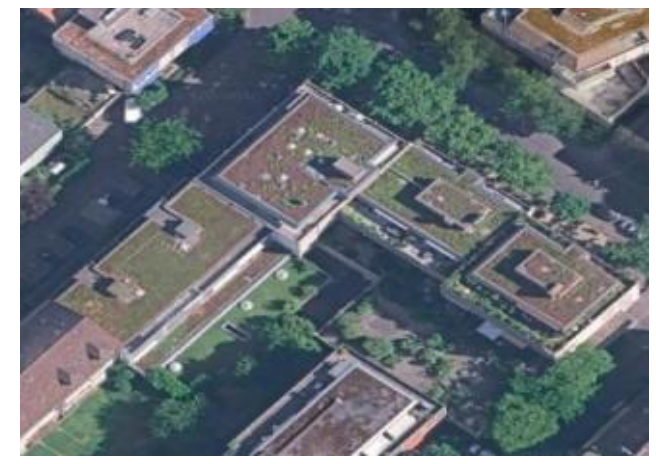 | The correct label is 'Green', but it is classified as 'Potential 'Green'. The area percentage of 'Potential Green' is 50% and the area percentage of 'Green' class is 23%. |
| 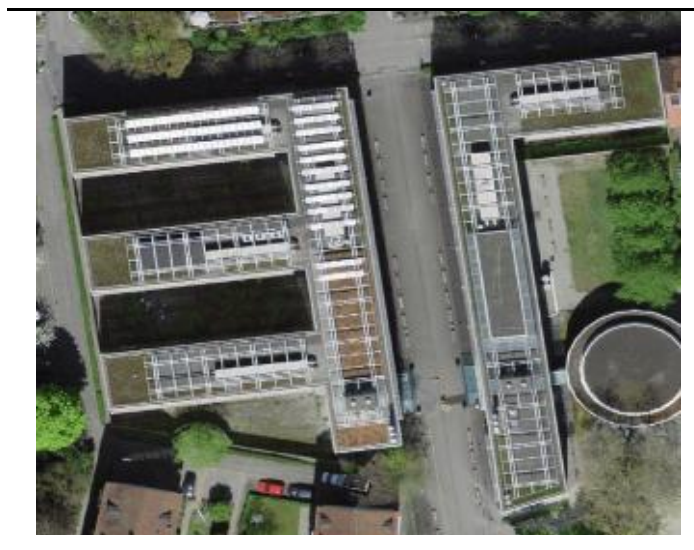 | Misclassified as 'Solar' due to the structure of the roof. |
| 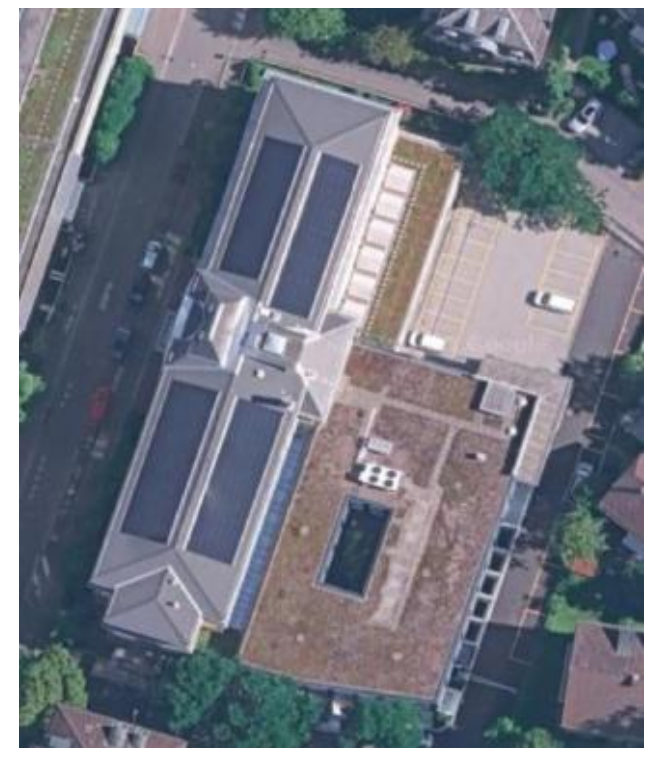 | Misclassified the building as 'Solar' instead of 'Green' due to the structure of the roof. The computed area percentage of 'Solar' class is 58%, 'Green' class is 9%, 'Potential Green' class is 25% and 'Not suitable' class is 8%. |

The second category arises from the area-based label assignment rather than from the segmentation model itself. In sample (b), the correct label is Green, but the building is assigned to Potential Green because the intersected area of the Potential Green prediction (50%) exceeds that of the Green prediction (23%). The pixel level prediction correctly identifies the vegetated portion of the roof, yet the building level label is overridden by the larger non-vegetated share of the footprint. This case illustrates the limitation discussed later in Section V: assigning a single dominant class to the whole footprint discards sub-building information and can demote a genuinely greened rooftop to a mere greening candidate.

The third category concerns geometric confusion with the Solar class. In samples (c) and (d), the regular linear structure of the roof, such as ribbed metal sheeting, skylight rows, or other repetitive elements, mimics the geometric signature of photovoltaic arrays, and the model assigns the Solar class. Sample (d) compounds both error types: the roof is structurally misread as Solar (58% of the footprint area) although the correct class is Green, whose prediction covers only 9%, with Potential Green at 25% and Not Suitable at 8%. Because the Solar share dominates, the area-based assignment propagates the segmentation error to the building level. These cases explain why the Solar class, although reliably detected overall, still contributes between 9.5 and 21.4% of the false positives across pipelines (TABLE 5): its detection relies on geometric regularity, which some non-solar roof structures share.

Taken together, the misclassified samples indicate two complementary directions for improvement. Confusions of the first and third type call for richer input features, such as the near-infrared band for separating living vegetation from vegetation-like materials, and higher resolution or elevation-derived texture for distinguishing panel arrays from structured roofing. Confusions of the second type call for a refinement of the label assignment stage, for example by reporting per-class coverage fractions alongside the dominant label, or by introducing class-specific priority rules for small but decisive classes such as Green.

## V. Discussion and Limitations

Although the average mIoU of the pipelines is high, many obvious mis-extractions of rooftops can still be found in the output results, as presented in Fig. 14. For the Roofpedia and Modified Roofpedia models, it is difficult to identify the exact extent of a rooftop when part of it is covered by overhanging trees or their shadows, and this inaccurate segmentation reduces the accuracy of the results. proj-vegroofs-DL addressed this issue by computing a Canopy Height Model (CHM) from LiDAR data and masking the pixels belonging to overhanging trees, an approach that could be adopted in future versions of the Modified Roofpedia workflow.

It is also evident that the outputs of the Modified Roofpedia model trained on SwissImage RS and of proj-vegroofs-DL do not reach the quality of the Roofpedia and the other Modified Roofpedia configuration. Notably, both of these models used the SwissImage RS dataset, which incorporates the near-infrared band. The SwissImage RS images are visibly blurry, which makes it difficult to determine sharp building boundaries. This suggests that the dataset is well suited to vegetation-oriented models, where the near-infrared signal outweighs the loss of geometric detail, but has clear limitations for building-scale models that depend on crisp edges, such as rooftop segmentation. Finally, while the mIoU curves provided a stable basis for comparing the pipelines, the loss values fluctuated considerably during validation, so loss alone is a less reliable indicator of rooftop extraction performance.

In summary, a modified Roofpedia deep learning architecture was proposed and evaluated against the original Roofpedia model and the deep learning rooftop segmentation model from the Swiss Territorial Data Lab (STDL). The improved model integrates multisource open geospatial datasets from Swisstopo, incorporates elevation and slope-based constraints derived from open-source elevation models, and outperforms the original Roofpedia model in identifying suitable rooftop surfaces. A systematic comparison between the original and modified predictions is provided for reproducibility.

The results further show that the labelling strategy significantly affects the distribution of rooftop classes, even when it is applied to identical deep learning predictions. The priority-based method tends to overweight solar rooftops, producing higher shares of the Solar (11.7%) and Potential Green (44.2%) classes. This approach is useful when policy frameworks explicitly prioritize solar energy expansion. In contrast, the area-based assignment produces a more physically grounded estimation, in which each rooftop is classified according to the dominant intersected area. This reduces the Solar share to 6.3% and increases the shares of the Green and Not Suitable classes correspondingly. The inclusion of the slope threshold (Average Slope > 6.5º) does not significantly alter the results for Bern, indicating that most buildings already meet the low-slope requirement for green roof suitability. Overall, these differences highlight the importance of selecting an appropriate labelling strategy according to the planning objective, whether the aim is to maximize solar adoption, identify green roof potential, or maintain a balanced representation of rooftop categories.

A limitation of the area-based label assignment is that the entire footprint of a building is counted as the covered area. In some cases, the predicted label covers only a small proportion of the rooftop, so assigning its class to the whole building over-attributes that class at the building level. Incorporating the true intersected area fraction or reporting sub-building coverage percentages alongside the building-level label, would mitigate this effect in future work.

## VI. Conclusion

Accurate rooftop classification is essential for supporting climate adaptation, urban greening, and sustainable infrastructure planning. This research bridges scientific evidence and real-world decision making by delivering high quality, policy relevant geospatial insights. By integrating robust remote sensing techniques with user centred analytical products, the study supports decision makers, stakeholders, scientists, and the wider public in understanding and responding to environmental change, contributing to more informed governance, improved climate resilience, and long-term sustainable development.

This study demonstrates that the modified Roofpedia model, built entirely on open-source Swiss geospatial datasets, can provide a detailed and scalable rooftop classification system for urban sustainability applications. The proposed four class formulation reached a validation mIoU of approximately 0.78 while delivering considerably richer output than the binary baselines of Roofpedia and proj-vegroofs-DL, since it separates existing green rooftops, potential green rooftops, solar rooftops, and unsuitable flat surfaces within a single model. The comparison of three aerial image sources showed that RGB imagery, whether from Mapbox or SwissImage, supports the full four class task, whereas the near infrared band of SwissImage RS benefits vegetation detection but limits building scale segmentation because of reduced image sharpness. Since Mapbox and SwissImage performed comparably but Mapbox restricts the number of tiles that can be retrieved through its API, SwissImage was adopted as the operational input for the city scale application. The evaluation of the two label assignment strategies further showed how methodological choices alone can shift building level class distributions, and therefore policy relevant metrics, even when the underlying predictions are identical. The workflow remains fully reproducible, and the results enhance the understanding of rooftop suitability for green infrastructure and contribute to evidence based urban climate adaptation planning.

Because all methodological components are built on open-source datasets, the framework can be readily transferred beyond Bern, enabling future applications across Switzerland and ultimately at the global scale. For Bern specifically, the resulting rooftop dataset directly supports evidence based urban planning, helping policymakers, architects, and environmental agencies design strategies for a cooler, more sustainable, and climate resilient city. Future work will extend the framework to additional Swiss and international cities, incorporate near infrared and LiDAR derived features such as canopy height models to reduce the remaining confusion between the Potential Green and Not Suitable classes and to handle overhanging trees, and couple the resulting rooftop datasets with urban climate simulations to quantify the cooling benefits of city wide green roof deployment. Such coupling is well motivated, since global assessments estimate that roof greening can lower daytime land surface temperatures by roughly 0.6 to 1.6 ℃ depending on the greening extent.

## Acknowledgment

Htet Y. K. K. gratefully acknowledges Swiss Territorial Data Lab for providing the SWISSIMAGE RS dataset for Bern study area. Htet Y. K. K. would like to express gratitude for the discussions with Mr. Ueli Mauch (Swisstopo), Frau Roxane Pott (Swisstopo), Prof. Dr. Michael Rast (ISSI), Prof. Dr. Stefan Brönnimann (Geographic Institute of University of Bern), Dr. Moritz Burger (Geographic Institute of University of Bern), and Prof. Dr. Edouard Davin (Wyss Academy for Nature, University of Bern) for providing expert knowledge and feedback of the model's results.

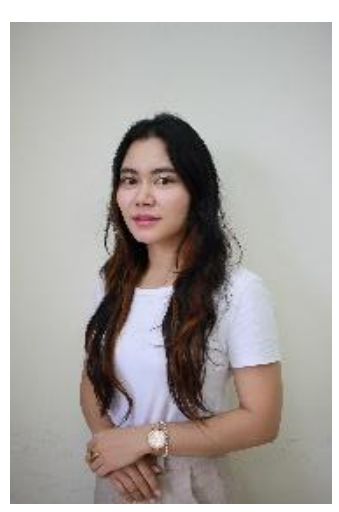

**Htet Y. K. K.** (Postdoctoral Fellow, ISSI) Htet Yamin Ko Ko received the Dr.Eng. degree in Remote Sensing and GIS from the Asian Institute of Technology (AIT), Thailand. She is currently a Postdoctoral Research Fellow at the International Space Science Institute (ISSI), Bern, Switzerland. Prior to joining ISSI, she worked as a Senior Research Associate at AIT Geoinformatics Center, and as a Research Assistant at King Mongkut’s University of Technology Thonburi (KMUTT), Thailand. Her research interests include satellite Earth observation, urban climate, geospatial deep learning, land surface temperature validation, local climate zone mapping, foundation models, and Nature-based solutions.